\documentclass[10pt,twocolumn,letterpaper]{article}

\usepackage{iccv}
\usepackage{times}
\usepackage{epsfig}
\usepackage{graphicx}
\usepackage{amsmath}
\usepackage{amssymb}

\usepackage{multirow}
\usepackage{array}
\usepackage{booktabs}
\usepackage{makecell}
\usepackage{mathtools}
\usepackage{afterpage}

\usepackage[accsupp]{axessibility}
\usepackage[pagebackref=true,breaklinks=true,letterpaper=true,colorlinks,bookmarks=false]{hyperref}

\iccvfinalcopy

\def\iccvPaperID{6582}

\ificcvfinal\fi

\makeatletter
\newcommand{\thickhline}{%
    \noalign {\ifnum 0=`}\fi \hrule height 1.2pt
    \futurelet \reserved@a \@xhline
}
\newcolumntype{"}{@{\hskip\tabcolsep\vrule width 1pt\hskip\tabcolsep}}
\makeatother

\begin{document}

\title{Estimating and Exploiting the Aleatoric Uncertainty \\ in Surface Normal Estimation}

\author{Gwangbin Bae \;\;\;\; Ignas Budvytis \;\;\;\; Roberto Cipolla\\
University of Cambridge\\
{\tt\small \{gb585,ib255,rc10001\}@cam.ac.uk}
}

\maketitle
\ificcvfinal\thispagestyle{empty}\fi

\begin{abstract}
Surface normal estimation from a single image is an important task in 3D scene understanding. In this paper, we address two limitations shared by the existing methods: the inability to estimate the aleatoric uncertainty and lack of detail in the prediction. The proposed network estimates the per-pixel surface normal probability distribution. We introduce a new parameterization for the distribution, such that its negative log-likelihood is the angular loss with learned attenuation. The expected value of the angular error is then used as a measure of the aleatoric uncertainty. We also present a novel decoder framework where pixel-wise multi-layer perceptrons are trained on a subset of pixels sampled based on the estimated uncertainty. The proposed uncertainty-guided sampling prevents the bias in training towards large planar surfaces and improves the quality of prediction, especially near object boundaries and on small structures. Experimental results show that the proposed method outperforms the state-of-the-art in ScanNet \cite{ScanNet} and NYUv2 \cite{NYUv2}, and that the estimated uncertainty correlates well with the prediction error. Code is available at \url{https://github.com/baegwangbin/surface_normal_uncertainty}.
\end{abstract}

\section{Introduction}
\label{sec:intro}

\begin{figure}[t]
\begin{center}
\includegraphics[width=1.0\linewidth]{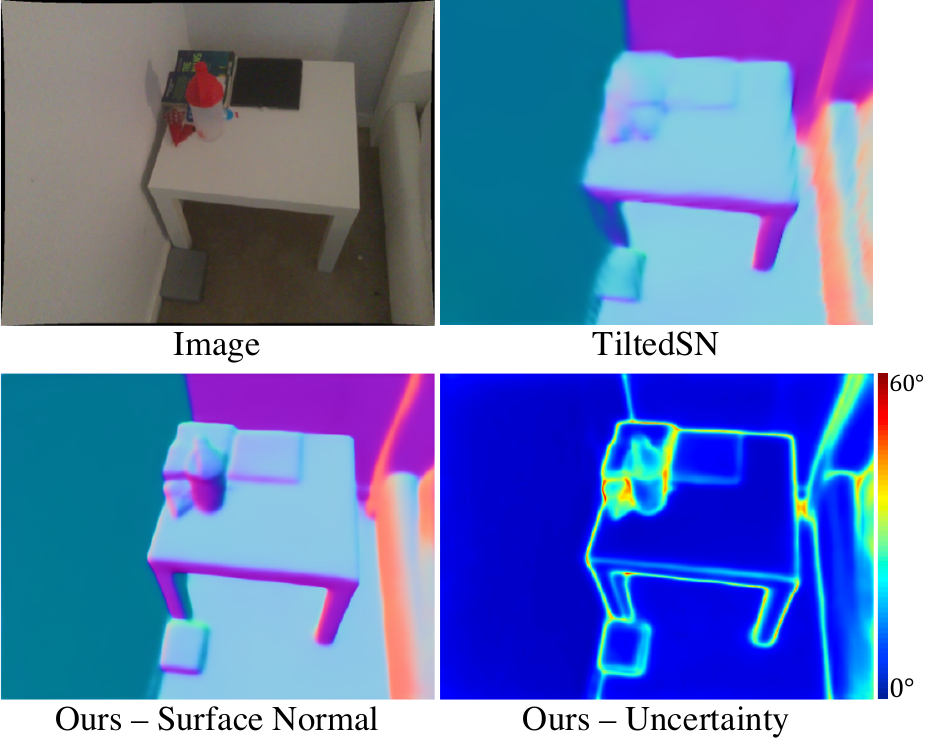}
\end{center}
\caption{Comparison between our method and TiltedSN \cite{SNfromRGB_20_TiltedSN}. The proposed network estimates the \textit{surface normal probability distribution}, from which the expected angular error can be inferred. The prediction made by our method shows clearer object boundaries and preserves a higher level of detail. This is due to the proposed uncertainty-guided sampling which prevents the bias in training towards large planar surfaces.}
\label{fig:intro}
\end{figure}

The ability to estimate surface normal from a single RGB image plays a crucial role in understanding the 3D scene geometry. The estimated normal can be used to build augmented reality (AR) applications \cite{SNfromRGB_19_FrameNet} or to control autonomous robots \cite{SN_robot2}. In this work, we address two limitations shared by the state-of-the-art methods.

\begin{figure}[t]
\begin{center}
\includegraphics[width=1.0\linewidth]{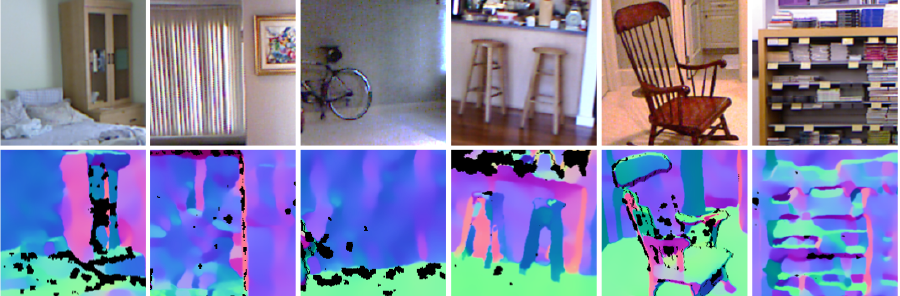}
\end{center}
\caption{Ground truth surface normal of NYUv2 \cite{NYUv2}, generated by Ladicky et al. \cite{SNfromRGB_14_Ladicky}. The ground truth is unreliable especially near object boundaries and on small structures.}
\label{fig:motivation}
\end{figure}

\textit{(1) Inability to estimate the aleatoric uncertainty.} State-of-the-art learning-based approaches \cite{SNfromRGB_15_Deep3D,SNfromRGB_15_Eigen,SNfromRGB_16_SkipNet,SNfromRGB_18_GeoNet,SNfromRGB_19_FloorsAreFlat,SNfromRGB_19_FrameNet,SNfromRGB_19_PAP,SNfromRGB_19_SR,SNfromRGB_20_GeoNet++,SNfromRGB_20_TiltedSN,SNfromRGB_20_VPLNet} train deep networks by minimizing some distance metric (e.g., $L_2$) between the predicted normal and the ground truth. However, the ground truth normal, calculated from a measured depth map, can be sensitive to the depth noise and to the algorithm used to compute the normal (see Fig. \ref{fig:motivation} for examples of inaccurate ground truth). The network should be able to capture such aleatoric uncertainty, in order to be deployed in real-world applications.

\textit{(2) Lack of detail in the prediction.} An indoor scene generally consists of large planar surfaces (e.g., walls and floors) and small objects with complex geometry. Therefore, if the training loss is applied to all pixels, the learning becomes biased to large surfaces, resulting in an over-smoothed output. Such bias can be solved by applying the loss on a carefully selected \textit{subset} of pixels. For example, in \cite{ranking_loss_depth}, pair-wise ranking loss was applied to the pixels near instance boundaries to improve the quality of monocular depth estimation. However, such effort has not been made for surface normal estimation.

In this work, we estimate the aleatoric uncertainty by predicting the \textit{probability distribution} of the per-pixel surface normal. While the von Mises-Fisher distribution \cite{fisher1993statistical} can be used for this purpose, minimizing its negative log-likelihood (NLL) is equivalent to minimizing the $L_2$ distance between the predicted normal and the ground truth with learned loss attenuation. As the error metric of our interest is the \textit{angle} between the two vectors, we introduce a new parameterization for the distribution such that its NLL is the angular loss with learned attenuation. At test time, the expected angular error is calculated from the estimated distribution, and used as a measure of the aleatoric uncertainty.

We also propose a novel decoder framework to improve the level of detail in the prediction. The network initially makes a coarse prediction for which the training loss is applied to all pixels. Then, the coarse prediction and feature-map are bilinearly upsampled by a factor of 2, and are passed through a pixel-wise multi-layer perceptron (MLP) to yield a refined output. This process is repeated until reaching the desired resolution. The MLPs are trained on a subset of pixels selected based on the uncertainty: Pixels with the highest uncertainty are selected and are complemented with uniformly sampled pixels. Such uncertainty-guided sampling prevents the bias in training towards large planar surfaces (for which the network estimates low uncertainty), thereby improving the quality of prediction near object boundaries and on small structures. 

Our contributions can be summarized as follows:

\begin{itemize}
    \item \textbf{Estimation of the aleatoric uncertainty in surface normal.} To the best of our knowledge, we are the first to estimate the aleatoric uncertainty in CNN-based surface normal estimation. We introduce a new parameterization for the surface normal probability distribution and show that the estimated uncertainty correlates well with the prediction error.
    
    \item \textbf{Uncertainty-guided sampling for pixel-wise refinement.} We introduce a novel decoder module where the loss is applied to a subset of pixels selected based on the uncertainty. We show that this module significantly improves the quantitative and qualitative performance.

    \item \textbf{State-of-the-art performance.} Experimental results show that the proposed method achieves state-of-the-art performance on ScanNet \cite{ScanNet} and NYUv2 \cite{NYUv2}. Qualitatively, the prediction made by our method contains a higher level of detail (see Fig. \ref{fig:intro}).
\end{itemize}

\section{Related Work}
\label{sec:related_work}

\noindent
\textbf{Surface normal estimation.} Surface normal estimation from a single RGB image has been studied extensively in literature \cite{SNfromRGB_13_3DP,SNfromRGB_14_Fouhey,SNfromRGB_14_Ladicky,SNfromRGB_15_Deep3D,SNfromRGB_15_Eigen,SNfromRGB_16_SkipNet,SNfromRGB_16_SURGE,SNfromRGB_18_GeoNet,SNfromRGB_19_FrameNet,SNfromRGB_19_PAP,SNfromRGB_19_SR,SNfromRGB_20_GeoNet++,SNfromRGB_20_TiltedSN,SNfromRGB_20_VPLNet}. The existing methods generally consist of a feature extractor followed by a prediction head. For example, Ladicky et al. \cite{SNfromRGB_14_Ladicky} extracted hand-crafted features (e.g., SIFT \cite{sift}) and applied multi-class Ada-boost \cite{adaboost} to regress the output as a linear combination of a discrete set of normals. Following the success of deep learning, recent methods replace both components with convolutional neural networks (CNNs). 

Wang et al. \cite{SNfromRGB_15_Deep3D} introduced two-stream CNNs to learn global and local cues, and fused them with another CNN. Eigen and Fergus \cite{SNfromRGB_15_Eigen} proposed a multi-scale architecture to jointly predict depth, surface normals and semantic labels. Following these early attempts, contributions have been made by enforcing depth-normal consistency \cite{SNfromRGB_18_GeoNet,SNfromRGB_20_GeoNet++}, formulating the task as spherical regression \cite{SNfromRGB_19_SR}, and introducing a spatial rectifier to handle tilted images \cite{SNfromRGB_20_TiltedSN}. In this work, we address the aleatoric uncertainty in surface normal, which has not been studied in previous literature.

\noindent
\textbf{Uncertainty in deep learning.} Two major types of uncertainty are epistemic and aleatoric \cite{aleatory_or_epistemic}. Epistemic uncertainty (i.e. uncertainty in model) can be modelled by approximating the posterior over the model weights. For example, by applying dropout \cite{dropout} at test time, $N$ networks can be sampled from the approximate posterior, and the variance of the outputs can be used as a measure of uncertainty \cite{epiU-drop-gal}. The posterior can also be approximated by training $N$ networks on random subsets of data \cite{epiU-2017-boot}, or by taking $N$ snapshots during a single training \cite{epiU-2017-snap}. The aforementioned approaches are task-independent and can easily be applied to surface normal estimation.

The focus of this paper is on the aleatoric uncertainty, which captures the noise inherent in the data. We assume that the uncertainty is heteroscedastic \cite{what_uncertainties} (i.e. certain pixels have higher uncertainty than the others). For such a scenario, a commonly used approach is to estimate the per-pixel probability distribution over the output, and train the network by maximizing the likelihood of the ground truth \cite{what_uncertainties,prob_deep_net}. This requires a task-specific formulation and has not been studied for CNN-based surface normal estimation.

\noindent
\textbf{Distribution on a unit sphere.} The surface normal probability distribution should be defined on a unit sphere. An example of such distribution is the von Mises-Fisher distribution \cite{fisher1993statistical}, a rotationally symmetric uni-modal distribution defined on an $n$-sphere. In this paper, we introduce a variant of the von Mises-Fisher distribution, such that minimizing its negative log-likelihood is equivalent to minimizing the angle between the predicted normal and the ground truth, which is the error metric of our interest.

\noindent
\textbf{Uncertainty-guided sampling.} PointRend \cite{pointrend} is a neural network module designed for instance/semantic segmentation. As making inference on a regular grid leads to under-sampling of the pixels near object boundaries, PointRend uses a point-wise MLP to make inference on a subset of pixels with high uncertainty. Our decoder module is a novel extension of such a framework to surface normal estimation.

\section{Method}
\label{sec:method}

This section provides the details of our method. Firstly, we introduce a new parameterization for the surface normal probability distribution that can be used for uncertainty estimation. Secondly, we explain the network architecture and the uncertainty-guided sampling used for training the pixel-wise refinement networks.

\subsection{Aleatoric Uncertainty in Surface Normal}
\label{sec:method1}

Our goal is to learn the per-pixel surface normal probability distribution $p_i(\mathbf{n}_i|\mathcal{I})$, where $i$ is the pixel index and $\mathcal{I}$ is the input image. In practice, we parameterize the distribution with a set of parameters $\boldsymbol{\theta}_i$, which is estimated by a network of weights $\mathbf{W}$. The network is trained by minimizing the negative log-likelihood (NLL) of the ground truth $\mathbf{n}_i^\text{gt}$. The training loss can thus be written as
\begin{align}
    \label{eqn:aleatoric_loss}
    \mathcal{L} = - \frac{1}{N} \sum_i \log p_i(\mathbf{n}^\text{gt}_i|
    \boldsymbol{\theta}_i(\mathcal{I},\mathbf{W})
    ),
\end{align}
\noindent
where $N$ is the number of pixels with ground truth. Finding a suitable parameterization for the distribution is important as it determines which quantity will be minimized (or maximized) during training.

\begin{figure}[t]
\begin{center}
\includegraphics[width=1.0\linewidth]{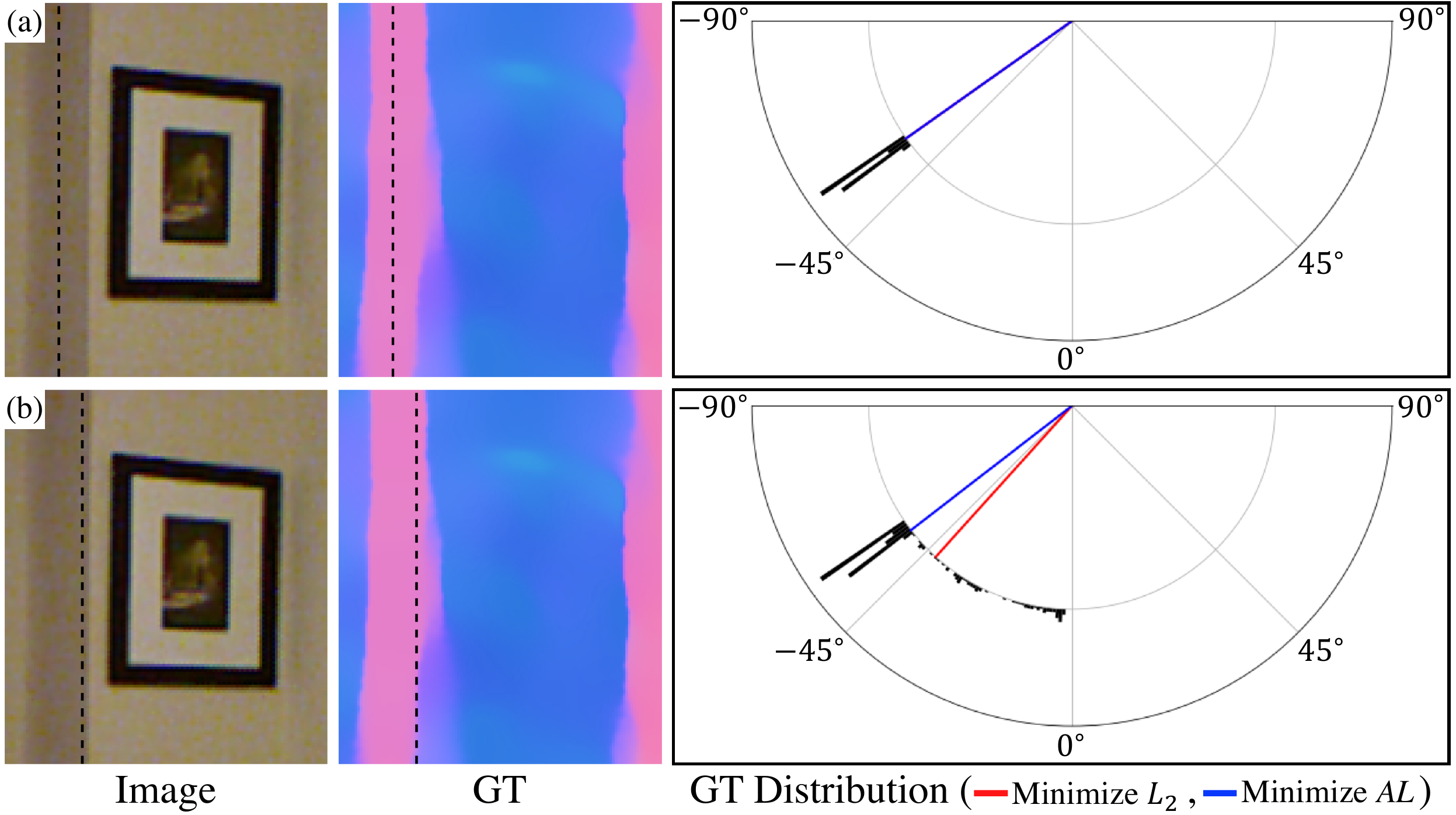}
\end{center}
\caption{Each histogram shows the distribution of ground truth along the dashed line. The red and the blue lines show the direction that minimizes the $L_2$ loss and the angular loss, respectively (the lines overlap for (a)). In both examples, the pixels along the dashed line have similar visual features and belong to the same plane. However, the pixels in (b) suffer from the noise caused by the neighboring pixels belonging to a different plane. The angular loss is more robust in the presence of such asymmetric noise.}
\label{fig:robustness_of_AL}
\end{figure}

\begin{figure*}[t]
\begin{center}
\includegraphics[width=1.0\linewidth]{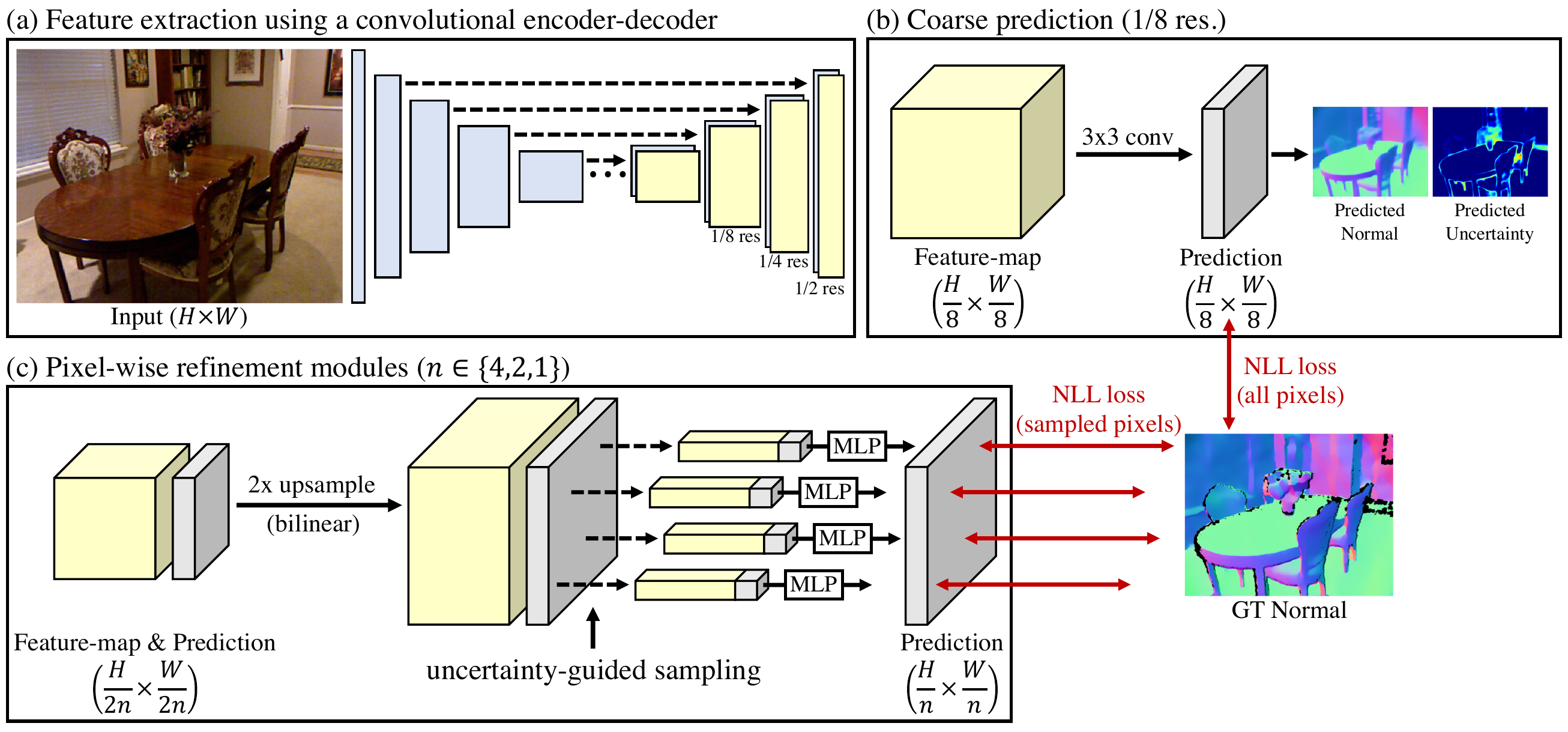}
\end{center}
\caption{Illustration of the proposed pipeline. Initially, a coarse prediction is made from the 1/8 resolution feature-map and the loss is applied to all pixels. Then, a refinement module upsamples the coarse feature-map and prediction by a factor of 2, and applies a pixel-wise MLP to yield a refined, higher resolution output. Full-resolution output is obtained by applying three refinement modules. The MLPs are trained on a subset of pixels selected based on the uncertainty, to prevent the bias in training towards low-uncertainty pixels.}
\label{fig:method}
\end{figure*}

\noindent
\textbf{von Mises-Fisher distribution.} We use the von Mises-Fisher distribution \cite{fisher1993statistical} (abbreviated hereafter as vonMF) as a baseline. It is a spherical analogue to the normal distribution, defined on a unit $n$-sphere in $\mathbb{R}^{n+1}$ \cite{moments_of_vMF}. For $n=2$, the probability density function (PDF) is given as
\begin{align}
    \label{eqn:vonMF-PDF}
    p_{\text{vonMF},i}(\mathbf{n}_i|\boldsymbol{\mu}_i,\kappa_i) = \frac{\kappa_i \exp(\kappa_i \boldsymbol{\mu}_i^T \mathbf{n}_i)}{4\pi \sinh \kappa_i},
\end{align}
\noindent
where $\boldsymbol{\mu}_i$ is the mean direction and $\kappa_i$ is the concentration parameter. Both $\mathbf{n}_i$ and $\boldsymbol{\mu}_i$ are unit vectors and $\kappa_i \geq 0$. Higher value of $\kappa_i$ means that the distribution is more concentrated around $\boldsymbol{\mu}_i$ and that the uncertainty is low for that pixel (the distribution is uniform when $\kappa_i=0$). The pixel-wise NLL loss can be written as
\begin{equation}
\label{eqn:vonMF-NLL}
\mathcal{L}_{\text{vonMF,}i} = -\log \kappa_{i} + \log \sinh{\kappa_i} - \kappa_i \boldsymbol{\mu}^T_{i} \mathbf{n}^{\text{gt}}_{i}.
\end{equation}

Maximizing $\boldsymbol{\mu}^T_{i} \mathbf{n}^{gt}_{i}$ is equivalent to minimizing the $L_2$ distance $|\boldsymbol{\mu}_{i} - \mathbf{n}^{gt}_{i}|^2_2$. The loss is attenuated for the pixels with high uncertainty. The first two terms in Eq. \ref{eqn:vonMF-NLL} prevent the network from predicting infinite $\kappa$ for all pixels. To summarize, Eq. \ref{eqn:vonMF-NLL} is an $L_2$ loss with learned attenuation.


\noindent
\textbf{\textit{Angular} vonMF distribution.} While Eq. \ref{eqn:vonMF-NLL} minimizes $L_2$, we argue that the loss should minimize the \textit{angle} between the predicted normal and the ground truth, $\cos^{-1}{\boldsymbol{\mu}^T_{i} \mathbf{n}^{gt}_{i}}$. Firstly, this makes the loss consistent with the error metric. Secondly, this makes the network more robust against the \textit{asymmetric} noise in the ground truth surface normal. 

The ground truth surface normal of a pixel is obtained by fitting a plane to the point cloud defined by the pixel and its local neighborhood. If some of the neighboring pixels belong to a different plane (e.g., because the central pixel is close to the plane boundary), the ground truth will be affected accordingly and the noise in the ground truth will be asymmetric around the true normal. The mean direction, which minimizes the $L_2$ loss, is sensitive to such asymmetric noise. The angular loss, on the other hand, is minimized at the median direction, which is more robust against such noise (see Fig. \ref{fig:robustness_of_AL}). To this end, we introduce a distribution such that its NLL is the angular loss with learned attenuation. The PDF and the NLL loss are given as,

\begin{align}
\label{eqn:ang-PDF}
p_{\text{AngMF},i}(\mathbf{n}_i|\boldsymbol{\mu}_i,\kappa_i) = 
\frac{(\kappa_i^2+1)\exp(-\kappa_i \cos^{-1} \boldsymbol{\mu}_i^T \mathbf{n}_i)}{2\pi (1 + \exp(-\kappa_i \pi))}
\end{align}
\begin{multline}
\label{eqn:ang-NLL}
\text{and} \;\;\; \mathcal{L}_{\text{AngMF},i} = -\log (\kappa^2_{i} +1) 
+\log (1 + \exp(-\kappa_i \pi)) \\
+\kappa_i \cos^{-1} \boldsymbol{\mu}_i^T \mathbf{n}_i^\text{gt}.
\end{multline}

We call this the \textit{Angular} vonMF distribution (abbreviated hereafter as AngMF). Eq. \ref{eqn:ang-PDF} is obtained by setting the NLL as $\mathcal{L}_i = C(\kappa_i) + \kappa_i \cos^{-1} \boldsymbol{\mu}_i^T \mathbf{n}_i$ and finding the expression for $C(\kappa_i)$ via normalization (derivation in the supplementary material). Minimizing Eq. \ref{eqn:ang-NLL} is equivalent to minimizing the angular error, while attenuating the loss for the pixels with high uncertainty (i.e. low $\kappa$). We show in the experiments that using the proposed AngMF leads to higher accuracy than using the vonMF.

\noindent
\textbf{Measure of uncertainty.} In the proposed distribution (Eq. \ref{eqn:ang-PDF}), $\kappa_i$ encodes the network's confidence in the predicted mean $\boldsymbol{\mu}_i$. To translate this into an intuitive quantity, we calculate the \textit{expected value} of the angular error

\begin{align}
\label{eqn:ang-uncertainty}
E[\cos^{-1} \boldsymbol{\mu}_i^T \mathbf{n}_i] 
= \frac{2\kappa_i}{\kappa_i^2+1}
+ \frac{\exp(-\kappa_i \pi)\pi}{1 + \exp(-\kappa_i \pi)},
\end{align}

\noindent
and use it as a measure of the pixel-wise aleatoric uncertainty (derivation in the supplementary material).

\subsection{Uncertainty-Guided Sampling for Pixel-Wise Refinement}
\label{sec:method2}

The NLL losses (Eq. \ref{eqn:vonMF-NLL} and Eq. \ref{eqn:ang-NLL}) are more robust against noisy data than their counterparts ($L_2$ and angular loss) as the loss is attenuated for high-uncertainty pixels. However, this also makes the training more biased to large planar surfaces that have low surface normal uncertainty. Such bias leads to the lack of detail in the prediction, as the network is not encouraged to make accurate predictions for the challenging pixels, most of which are near object boundaries and on small structures. To this end, we propose a novel decoder framework, where pixel-wise multi-layer perceptrons (MLPs) are trained on a subset of pixels selected based on the estimated uncertainty.

\noindent
\textbf{Feature extraction.} The proposed pipeline is illustrated in Fig. \ref{fig:method}. The input to the network is an RGB image of size $(H \times W)$. We first generate feature-maps of different resolutions, using a convolutional encoder-decoder with skip-connections. We use the same architecture as \cite{adabins}.

\noindent
\textbf{Coarse prediction.} The network initially makes a coarse prediction from the 1/8 resolution feature-map, using a $3\times 3$ convolutional layer. The number of output channels is 4 (3 for $\boldsymbol{\mu}$ and 1 for $\kappa$). The first three channels are $L_2$-normalized to ensure $||\boldsymbol{\mu}||=1$. We apply the modified ELU function \cite{other-ELU}, $f(x)=\text{ELU}(x)+1$, for the last channel to ensure that $\kappa$ is positive. For the coarse prediction, the training loss (Eq. \ref{eqn:ang-NLL}) is applied to \textit{all} pixels.

\noindent
\textbf{Pixel-wise refinement modules.} The coarse prediction is then passed through three pixel-wise refinement modules of the same architecture. The input to each module is a low-resolution feature-map and prediction of size $(H/2n \times  W/2n)$, and the output is a refined prediction of size $(H/n \times  W/n)$. The forward pass in each module consists of three steps. \textit{(1) Upsampling:} Both the feature-map and prediction are bilinearly upsampled by a factor of 2. \textit{(2) Uncertainty-guided sampling:} During training, a subset of pixels is selected base on the uncertainty. The sampling strategy is explained below in more detail. \textit{(3) Pixel-wise refinement:} An MLP with three hidden layers, each with 128 nodes and a ReLU \cite{other-RELU} activation, estimates a refined output for the sampled pixels. The input to the MLP is a concatenated vector of the pixel-wise feature and prediction. Same as in the coarse prediction layer, $L_2$ normalization and the modified ELU activation are applied to $\boldsymbol{\mu}$ and $\kappa$. During training, the loss is calculated only for the sampled pixels. At test time, the trained MLPs are applied to all pixels.

\begin{figure}[t]
\begin{center}
\includegraphics[width=1.0\linewidth]{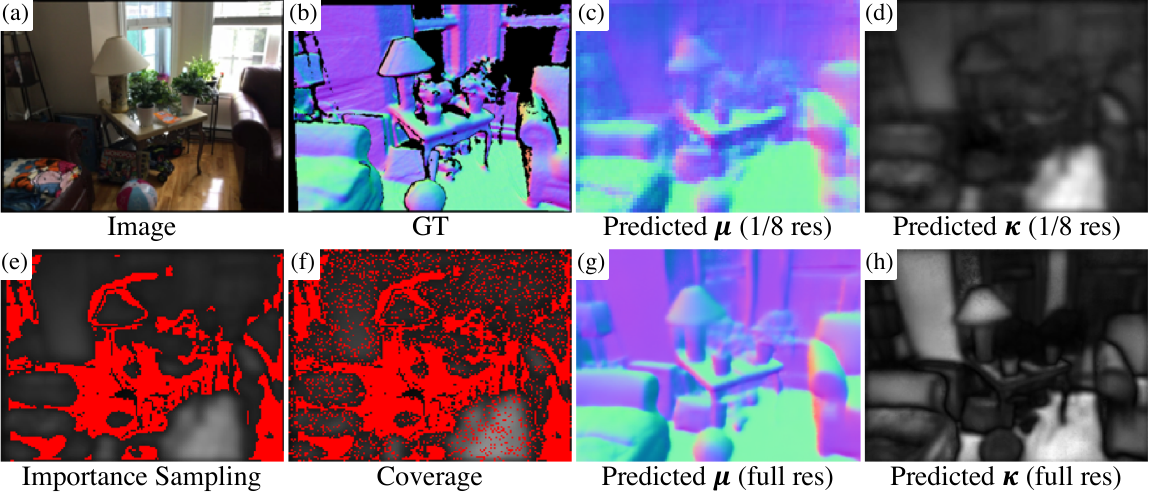}
\end{center}
\caption{(a-b) Input image and the ground truth. (c-d) Prediction made in coarse resolution, during the first epoch. In (d) and (h), white means high $\kappa$. The network estimates low $\kappa$ (i.e. high uncertainty) for most pixels except for those on the floor. If the NLL loss is applied to all pixels, the pixels on the floor will dominate the training as our loss is weighted by $\kappa$. (e-f) Uncertainty-guided sampling. We sample the pixels with high uncertainty (importance sampling) and add uniformly sampled pixels (coverage). Such sampling helps the network to focus on the challenging pixels. (g-h) Prediction made in full-resolution in the final epoch. The prediction is improved especially on the challenging pixels near object boundaries and on small structures. The network also becomes more confident about such pixels.}
\label{fig:sampling}
\end{figure}

\noindent
\textbf{Uncertainty-guided sampling.} Suppose that there are $h \cdot w$ pixels in the bilinearly upsampled prediction. In total, we sample $N_s = r_s \cdot h \cdot w$ pixels, where $r_s$ is set to $0.4$ in all experiments. Firstly, we sample $\beta_\text{UG} \cdot N_s$ pixels with the highest uncertainty (i.e. importance sampling). Then, $(1-\beta_\text{UG}) \cdot N_s$ pixels are sampled uniformly from the remaining pixels (i.e. coverage). $\beta_\text{UG}$, which can have values from 0 to 1, determines how biased the sampling is towards the high-uncertainty pixels. Fig. \ref{fig:sampling} illustrates the sampling process. 

\section{Experimental Setup}

\noindent
\textbf{Datasets.} We evaluate our method on two datasets: ScanNet \cite{ScanNet} and NYUv2 \cite{NYUv2}. ScanNet contains RGB-D frames from 1613 scans acquired in 807 different scenes. We use the ground truth surface normal and data split provided by FrameNet \cite{SNfromRGB_19_FrameNet}. NYUv2 consists of RGB-D video sequences capturing 464 indoor scenes. We evaluate on the official test set using the ground truth generated by Ladicky et al. \cite{SNfromRGB_14_Ladicky}. As the official training set only contains 795 images, state-of-the-art methods sample additional images from the training sequences \cite{SNfromRGB_15_Deep3D,SNfromRGB_16_SkipNet,SNfromRGB_18_GeoNet,SNfromRGB_20_GeoNet++} or supplement with other datasets \cite{SNfromRGB_19_SR,SNfromRGB_19_FloorsAreFlat}. To ensure a fair comparison, we use the same training set as GeoNet++ \cite{SNfromRGB_20_GeoNet++}.

\noindent
\textbf{Surface normal accuracy metrics.} Angular error is measured for the pixels with valid ground truth. Following \cite{SNfromRGB_13_3DP}, we report the mean, median and root-mean-squared error (lower the better), and the percentage of pixels with error below thresholds $t \! \in \! [11.25^{\circ}, 22.5^{\circ}, 30^{\circ}]$ (higher the better).

\noindent
\textbf{Uncertainty metrics.} The significance of the estimated uncertainty can be evaluated using sparsification curves \cite{on_the_uncertainty}. The pixels are sorted based on the uncertainty and an error metric $\epsilon$ is evaluated on the top $x$\% of pixels with low uncertainty. Following \cite{on_the_uncertainty}, we transform the accuracy metric (\% of pixels with error less than $t^\circ$) into an error metric by subtracting it from 100\%. We vary $x$ from 1 to 100, incrementing by 1, and report the area under the sparsification curve (AUSC) as in \cite{ausc_defined}. AUSC is affected by two factors: how accurate the predictions are, and how similar the uncertainty-based sorting is to the actual error-based sorting. To only evaluate the latter, we also report the area under the sparsification error (AUSE) \cite{ause_defined}, by subtracting the oracle sparsification (obtained via error-based sorting) from the estimated sparsification. 

\noindent
\textbf{Implementation details.} The proposed network is implemented with PyTorch \cite{PyTorch}. For training, we use the AdamW optimizer \cite{AdamW_introduced} and schedule the learning rate using 1cycle policy \cite{1cycle-lr} with $lr_\text{max} \! = \! 3.5 \! \times \! 10^{-4}$ (other hyper-parameters are set as their default values). The batch size is 4 and the number of epochs is 5 unless specified otherwise. 

\section{Experiments}

Firstly, we perform a set of ablation studies to demonstrate the effectiveness of the proposed approach. Then, the accuracy is compared against the state-of-the-art methods. Lastly, we evaluate the quality of the estimated uncertainty and compare it against alternative methods of uncertainty estimation.

\begin{table*}[t]
\setlength\tabcolsep{1.5pt}
\begin{center}
\begin{tabular}{l|c|ccc|ccccc}
\hline
\toprule
Architecture & Loss fn. & mean & median & rmse & $5.0^{\circ}$ & $7.5^{\circ}$ & $11.25^{\circ}$ & $22.5^{\circ}$ & $30^{\circ}$ \\
\midrule
\multirow{3}{*}{\small baseline} & {\small $L_2$}
& 13.53 & 7.22 & \textbf{21.16} & 35.10 & 51.44 & 65.08 & 82.38 & 87.83\\
\multirow{3}{*}{\small (convolutional encoder-decoder with skip connections \cite{adabins})} & {\small \textit{NLL-vonMF}} & 14.10 & 7.19 & 22.14 & 36.20 & 51.46 & 64.09 & 80.80 & 86.34\\
\cline{2-10}
& {\small \textit{AL}} & 13.45 & 6.70 & 21.78 & 38.65 & 54.04 & 66.73 & 82.46 & 87.53\\
& {\small \textit{NLL-AngMF}} & 13.82 & 6.60 & 22.47 & 39.69 & 54.30 & 65.97 & 81.64 & 86.71 \\
\hline
\hline
{\small baseline + pixel-wise MLPs} & \multirow{2}{*}{\small \textit{NLL-AngMF}} & 13.59 & 6.53 & 22.23 & 39.92 & 54.79 & 67.03 & 82.18 & 87.06 \\
{\small baseline + pixel-wise MLPs + uncertainty-guided sampling} & & 
\textbf{13.17} & \textbf{6.48} & 21.57 & \textbf{40.09} & \textbf{55.19} & \textbf{67.62} & \textbf{83.10} & \textbf{87.97} \\
\bottomrule
\end{tabular}
\end{center}
\caption{(top) The baseline network is trained with different loss functions. The proposed \textit{NLL-AngMF} shows higher accuracy than \textit{NLL-vonMF}, except for RMSE. \textit{NLL-AngMF} and \textit{NLL-vonMF} are \textit{AL} and $L_2$ with learned attenuation, respectively. As the training is biased to low-uncertainty pixels, the median error decreases, while RMSE increases. (bottom) The bias in training is solved by the proposed decoder modules. Both the pixel-wise MLPs and the uncertainty-guided sampling lead to improvement in all metrics.}
\label{table:ablation1}
\end{table*}

\subsection{Ablation Study}

The ablation study experiments are performed on a subset of ScanNet \cite{ScanNet}, obtained by sampling 20\% of the images in the training set (which contains 190K images).

\noindent
\textbf{Training loss.} \textit{NLL-vonMF} (Eq. \ref{eqn:vonMF-NLL}) is the $L_2$ loss with learned attenuation, and the proposed \textit{NLL-AngMF} (Eq. \ref{eqn:ang-NLL}) is the angular loss (\textit{AL}) with learned attenuation. We compare the four loss functions in Tab. \ref{table:ablation1} (top). As $L_2$ and \textit{AL} cannot be used for uncertainty estimation, the decoder modules are removed, and the surface normal is directly estimated from the convolutional encoder-decoder, by adding a $3\times 3$ convolutional layer to the final feature-map. Following are the key insights we can obtain from this experiment.

\begin{itemize}
    \item \textbf{\textit{NLL-AngMF} vs. \textit{NLL-vonMF}.} While \textit{NLL-vonMF} minimizes $L_2$, the proposed \textit{NLL-AngMF} minimizes the angular error, which is more consistent with the error metrics. As a result, \textit{NLL-AngMF} achieves significantly higher accuracy than \textit{NLL-vonMF}, except for RMSE.

    \item \textbf{\textit{NLL-AngMF} vs. \textit{AL}.} Our \textit{NLL-AngMF} is \textit{AL} with learned attenuation. As the training is biased to low-uncertainty pixels (mostly on large surfaces), the median error decreases and the accuracy for low thresholds ($5.0^{\circ}$ and $7.5^{\circ}$) increases. On the contrary, the mean error and RMSE increase and the accuracy for higher thresholds decreases. This is because the network is not penalized strongly for making inaccurate predictions for the challenging pixels.
\end{itemize}

\begin{figure}[t]
\begin{center}
\includegraphics[width=1.0\linewidth]{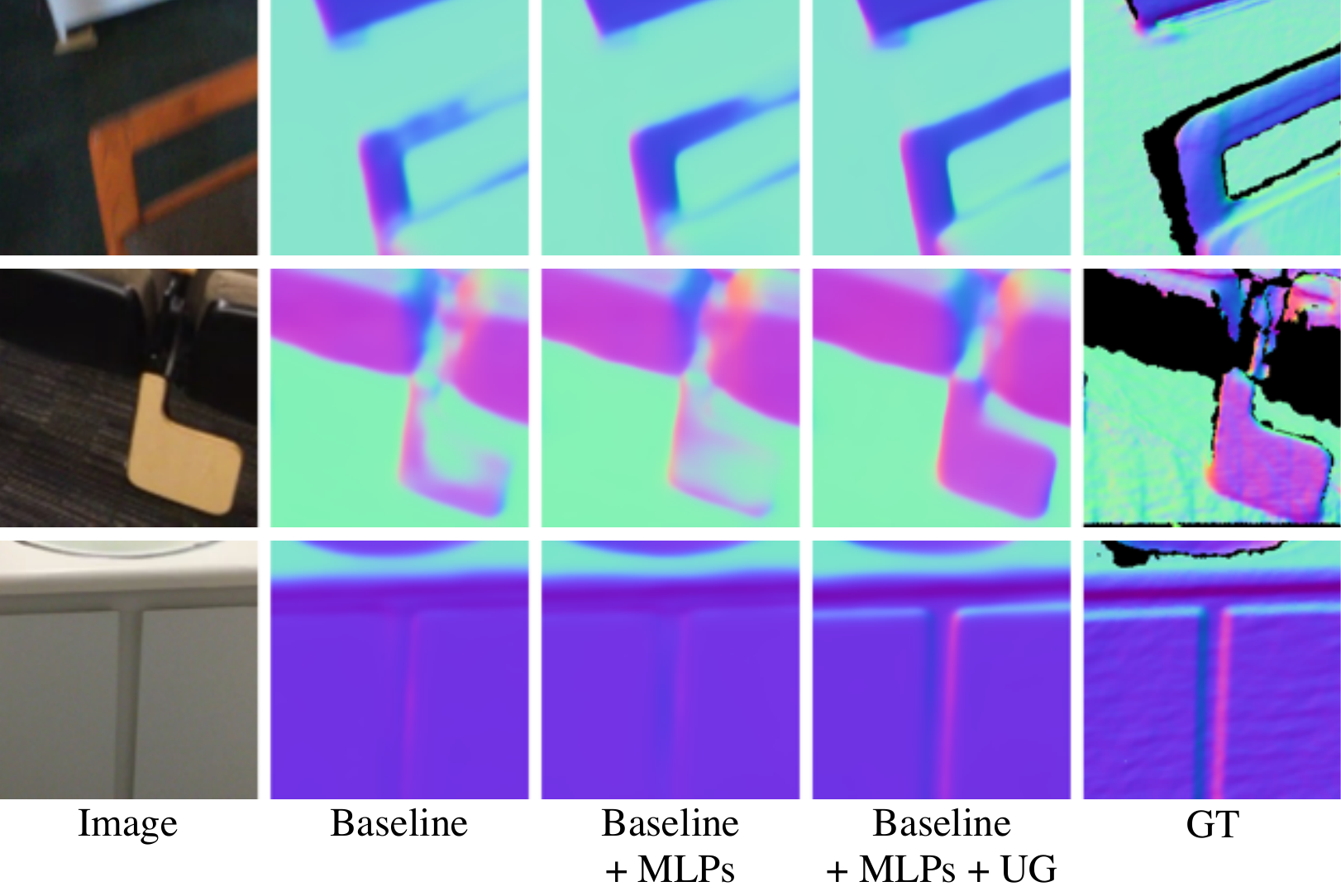}
\end{center}
\caption{Qualitative comparison between the networks with different decoder architecture (showing crops of 200 pixels $\times$ 200 pixels). The proposed uncertainty-guided sampling (UG) enforces the network to focus on the challenging pixels (i.e. those with high uncertainty). This improves the level of detail in the prediction.}
\label{fig:ablation1}
\end{figure}

\noindent
\textbf{Decoder architecture.} Tab. \ref{table:ablation1} (bottom) demonstrates the effectiveness of the proposed decoder modules. Firstly, we add the pixel-wise MLPs and train them on all pixels. Then, we apply the uncertainty-guided sampling during training (with $\beta_{\text{UG}}\!=\!0.7$). Both components lead to improvement in all metrics. As the uncertainty-guided sampling prevents the bias in training towards large planar surfaces, the quality of prediction is improved especially near object boundaries and on small structures, as shown in Fig. \ref{fig:ablation1}.

\noindent
\textbf{Sampling strategy.} Tab. \ref{table:ablation_beta} shows how the accuracy changes for different values of $\beta_{\text{UG}}$. $\beta_{\text{UG}}$ determines the ratio of the importance sampling. If $\beta_{\text{UG}}=1.0$, only the pixels with the highest uncertainty are sampled. If $\beta_{\text{UG}}=0.0$, the pixels are sampled uniformly. Finding the right balance between the two is important for minimizing the bias in training. Best performance is achieved when $\beta_\text{UG}=0.7$. 

\begin{table}[t]
\setlength\tabcolsep{1.5pt}
\begin{center}
\begin{tabular}{c|ccc|ccc}
\hline
\toprule
$\beta_{\text{UG}}$ & mean & median & rmse & $11.25^{\circ}$ & $22.5^{\circ}$ & $30^{\circ}$ \\
\midrule
0.0 & 13.58 & 6.52 & 22.18 & 66.68 & 82.09 & 87.09 \\
0.6 & 13.34 & 6.56 & 21.76 & 66.99 & 82.78 & 87.74 \\
0.7 & \textbf{13.17} & \textbf{6.48} & \textbf{21.57} & \textbf{67.62} & \textbf{83.10} & \textbf{87.97} \\
0.8 & 13.28 & 6.56 & 21.69 & 67.45 & 83.00 & 87.90 \\
1.0 & 13.26 & 6.59 & \textbf{21.57} & 67.16 & 82.98 & 87.92 \\
\bottomrule
\end{tabular}
\end{center}
\caption{Influence of $\beta_{\text{UG}}$ on the accuracy ($r_s$ is fixed to $0.4$). $\beta_{\text{UG}}$ is the ratio of the importance sampling. Best performance is achieved when $\beta_{\text{UG}}\!=\!0.7$.}
\label{table:ablation_beta}
\end{table}

\subsection{Comparison with the State-of-the-Art}

\begin{figure*}[t]
\begin{center}
\includegraphics[width=1.0\linewidth]{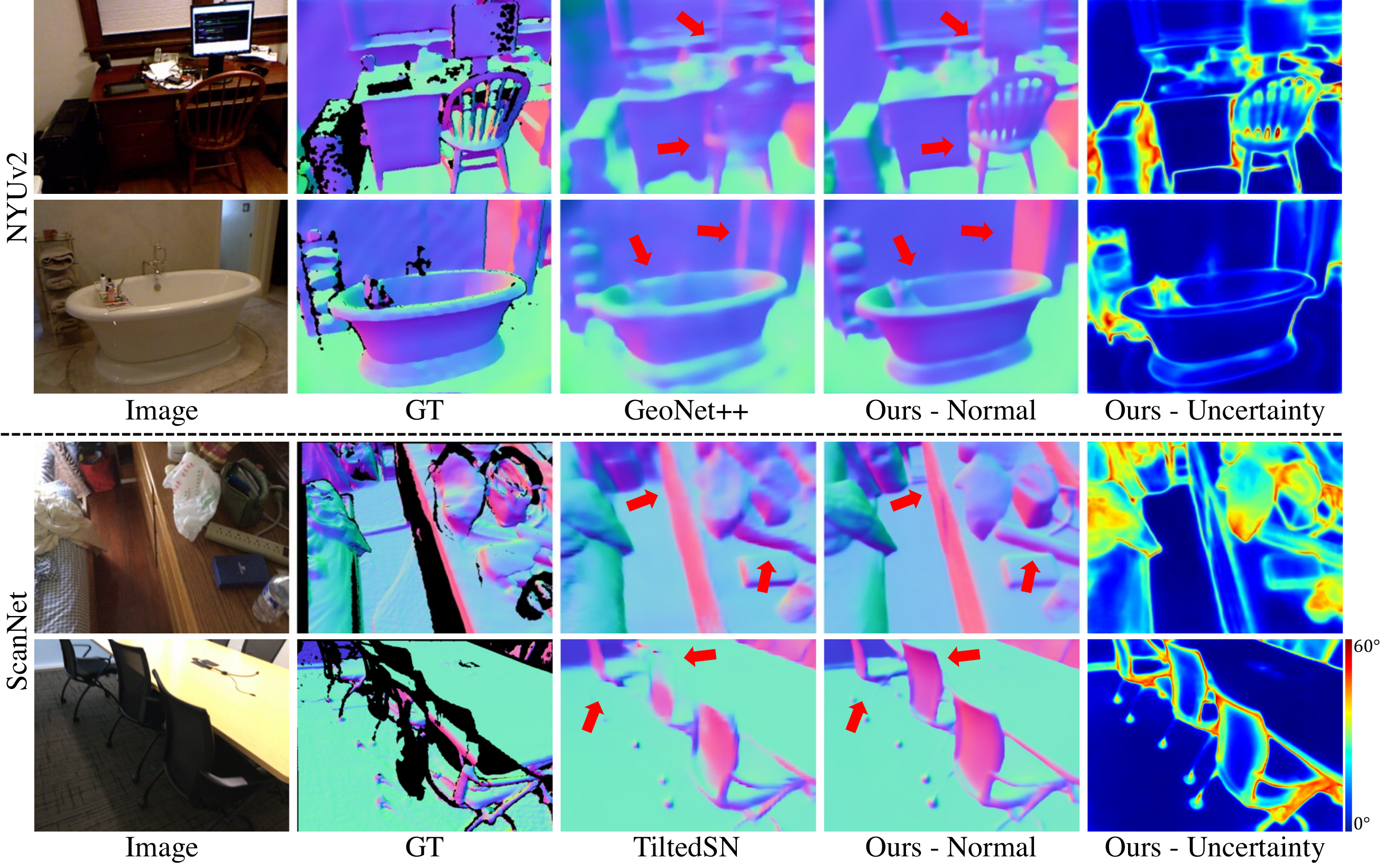}
\end{center}
\caption{Qualitative comparison against GeoNet++ \cite{SNfromRGB_20_GeoNet++} and TiltedSN \cite{SNfromRGB_20_TiltedSN}. The predictions made by our method show clearer object boundaries and preserve the fine-details of the scene geometry (see the regions pointed by the red arrows). The estimated uncertainty is high near object boundaries and on small structures. More examples are provided in the supplementary material.}
\label{fig:bm}
\end{figure*}

\begin{table}[t]
\small
\setlength\tabcolsep{1.5pt}
\begin{center}
\begin{tabular}{l|c|ccc|ccc}
\toprule
Method & Train & mean & median & rmse & $11.25^{\circ}$ & $22.5^{\circ}$ & $30^{\circ}$ \\
\midrule
Ladicky et al. \cite{SNfromRGB_14_Ladicky} & \multirow{11}{*}{N} & 33.5 & 23.1 & - & 27.5 & 49.0 & 58.7 \\
Fouhey et al. \cite{SNfromRGB_14_Fouhey} & & 35.2 & 17.9 & - & 40.5 & 54.1 & 58.9 \\
Deep3D \cite{SNfromRGB_15_Deep3D} & & 26.9 & 14.8 & - & 42.0 & 61.2 & 68.2 \\
Eigen et al. \cite{SNfromRGB_15_Eigen} & & 20.9 & 13.2 & - & 44.4 & 67.2 & 75.9 \\
SkipNet \cite{SNfromRGB_16_SkipNet} & & 19.8 & 12.0 & 28.2 & 47.9 & 70.0 & 77.8 \\
SURGE \cite{SNfromRGB_16_SURGE} & & 20.6 & 12.2 & - & 47.3 & 68.9 & 76.6 \\
GeoNet \cite{SNfromRGB_18_GeoNet} & & 19.0 & 11.8 & 26.9 & 48.4 & 71.5 & 79.5 \\
PAP \cite{SNfromRGB_19_PAP} & & 18.6 & 11.7 & 25.5 & 48.8 & 72.2 & 79.8 \\
GeoNet++ \cite{SNfromRGB_20_GeoNet++} & & 18.5 & 11.2 & 26.7 & 50.2 & 73.2 & 80.7 \\
\hline
Ours & N & \textbf{14.9} & \textbf{7.5} & \textbf{23.5} & \textbf{62.2} & \textbf{79.3} & \textbf{85.2} \\
\hline
\hline
FrameNet\cite{SNfromRGB_19_FrameNet} & & 18.6 & 11.0 & 26.8 & 50.7 & 72.0 & 79.5 \\
VPLNet\cite{SNfromRGB_20_VPLNet} & S & 18.0 & 9.8 & - & 54.3 & 73.8 & 80.7 \\
TiltedSN\cite{SNfromRGB_20_TiltedSN} & & 16.1 & \textbf{8.1} & 25.1 & \textbf{59.8} & 77.4 & 83.4 \\
\hline
Ours & S & \textbf{16.0} & 8.4 & \textbf{24.7} & 59.0 & \textbf{77.5} & \textbf{83.7} \\
\bottomrule
\end{tabular}
\end{center}
\caption{Surface normal accuracy on NYUv2 \cite{NYUv2}. The proposed method shows state-of-the-art performance. (top) The networks are trained on NYUv2. (bottom) The networks are trained on ScanNet \cite{ScanNet} and tested on NYUv2 without fine-tuning.}
\label{table:BM-nyu}
\end{table}

\begin{table}[t]
\setlength\tabcolsep{1.5pt}
\begin{center}
\begin{tabular}{l|ccc|ccc}
\toprule
Method  & mean & median & rmse & $11.25^{\circ}$ & $22.5^{\circ}$ & $30^{\circ}$ \\
\midrule
FrameNet\cite{SNfromRGB_19_FrameNet} & 14.7 & 7.7 & 22.8 & 62.5 & 80.1 & 85.8 \\
VPLNet\cite{SNfromRGB_20_VPLNet} & 13.8 & 6.7 & - & 66.3 & 81.8 & 87.0 \\
TiltedSN\cite{SNfromRGB_20_TiltedSN} & 12.6 & 6.0 & 21.1 & 69.3 & 83.9 & 88.6 \\
\hline
Ours & \textbf{11.8} & \textbf{5.7} & \textbf{20.0} & \textbf{71.1} & \textbf{85.4} & \textbf{89.8} \\
\bottomrule
\end{tabular}
\end{center}
\caption{Surface normal accuracy on ScanNet \cite{ScanNet}. Our method outperforms other methods across all metrics.}
\label{table:BM-scannet}
\end{table}

\noindent
\textbf{NYUv2.} Tab. \ref{table:BM-nyu} compares the accuracy of different methods on NYUv2 \cite{NYUv2}. Note that, compared to ScanNet \cite{ScanNet}, the quality of the ground truth is noticeably worse for NYUv2. While the ground truth for ScanNet is calculated from a 3D mesh that is obtained by fusing thousands of RGB-D frames, the ground truth for NYUv2 is calculated from a single noisy depth map. Nonetheless, the proposed training loss (angular loss with learned attenuation) and decoder framework (trained with uncertainty-guided sampling) help the network to learn from noisy data. As a result, our network shows a decisive improvement over GeoNet++ \cite{SNfromRGB_20_GeoNet++}. Qualitative comparison in Fig. \ref{fig:bm} shows that the predictions made by our method contain a higher level of detail. We also train the network on ScanNet and test on NYUv2 without fine-tuning. In this cross-dataset evaluation, we win against other methods except for the median error and $11.25^{\circ}$, suggesting that the network can generalize well to an unseen dataset.

\noindent
\textbf{ScanNet.} Tab. \ref{table:BM-scannet} compares different methods trained and tested on ScanNet \cite{ScanNet}. The batch size is set to 16 for this experiment. We outperform the state-of-the-art methods across all metrics. Qualitative comparison against TiltedSN \cite{SNfromRGB_20_TiltedSN} is provided in Fig. \ref{fig:bm}.

\subsection{Quality of Uncertainty}

\begin{figure}[t]
\begin{center}
\includegraphics[width=1.0\linewidth]{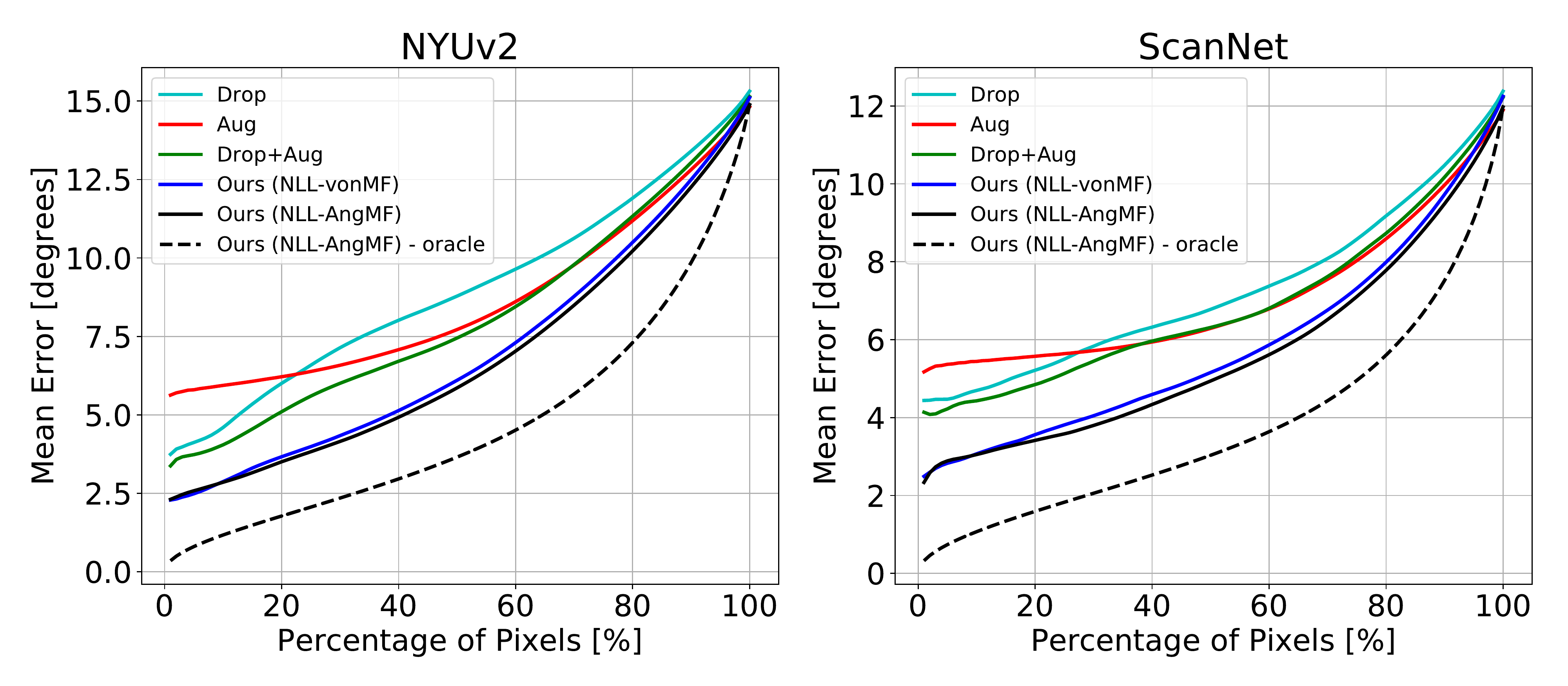}
\end{center}
\caption{Sparsification curves obtained by different methods of estimating the surface normal uncertainty.}
\label{fig:roc}
\end{figure}

\begin{table}[t]
\setlength\tabcolsep{1.5pt}
\begin{center}
\begin{tabular}{l|ccc|ccc}
\toprule
\multirow{2}{4em}{Method} & \multicolumn{3}{c|}{\small AUSC $\downarrow$} &
\multicolumn{3}{c}{\small AUSE $\downarrow$}\\
\cline{2-7}
& {\small mean} & {\small rmse} & {\footnotesize $11.25^{\circ}$}
& {\small mean} & {\small rmse} & {\footnotesize $11.25^{\circ}$}\\
\midrule
\textit{Drop} 
& 9.01 & 15.84 & 19.32 & 4.02 & 9.61 & 10.23\\
\textit{Aug} 
& 8.64 & 15.08 & 18.75 & 3.93 & 9.14 & 10.25\\
\textit{Drop + Aug} 
& 8.16 & 14.32 & 16.73 & 3.22 & 8.15 & 7.75\\
\hline
Ours \textit{(NLL-vonMF)} 
& 7.03 & 10.96 & 14.24 & \textbf{2.11} & \textbf{4.80} & 5.10\\
Ours \textit{(NLL-AngMF)} 
& \textbf{6.83} & \textbf{10.92} & \textbf{13.47} & 2.13 & 4.98 & \textbf{5.01}\\
\bottomrule
\end{tabular}
\end{center}
\caption{Quantitative evaluation of uncertainty on NYUv2 \cite{NYUv2}.}
\label{table:uncertainty-nyu}
\end{table}

\begin{table}[t]
\setlength\tabcolsep{1.5pt}
\begin{center}
\begin{tabular}{l|ccc|ccc}
\toprule
\multirow{2}{4em}{Method} & \multicolumn{3}{c|}{\small AUSC $\downarrow$} &
\multicolumn{3}{c}{\small AUSE $\downarrow$}\\
\cline{2-7}
& {\small mean} & {\small rmse} & {\footnotesize $11.25^{\circ}$}
& {\small mean} & {\small rmse} & {\footnotesize $11.25^{\circ}$}\\
\midrule
\textit{Drop} 
& 7.25 & 12.51 & 13.95 & 3.24 & 7.55 & 8.58\\
\textit{Aug} 
& 7.06 & 12.58 & 13.72 & 3.32 & 7.92 & 8.81\\
\textit{Drop + Aug} 
& 6.87 & 12.07 & 12.73 & 2.93 & 7.20 & 7.49\\
\hline
Ours \textit{(NLL-vonMF)} 
& 5.84 & 9.30 & 10.31 & \textbf{1.85} & \textbf{4.38} & 4.69\\
Ours \textit{(NLL-AngMF)} 
& \textbf{5.64} & \textbf{9.07} & \textbf{9.48} & 1.88 & \textbf{4.38} & \textbf{4.47}\\
\bottomrule
\end{tabular}
\end{center}
\caption{Quantitative evaluation of uncertainty on ScanNet \cite{ScanNet}.}
\label{table:uncertainty-scannet}
\end{table}

\begin{figure}[t]
\begin{center}
\includegraphics[width=1.0\linewidth]{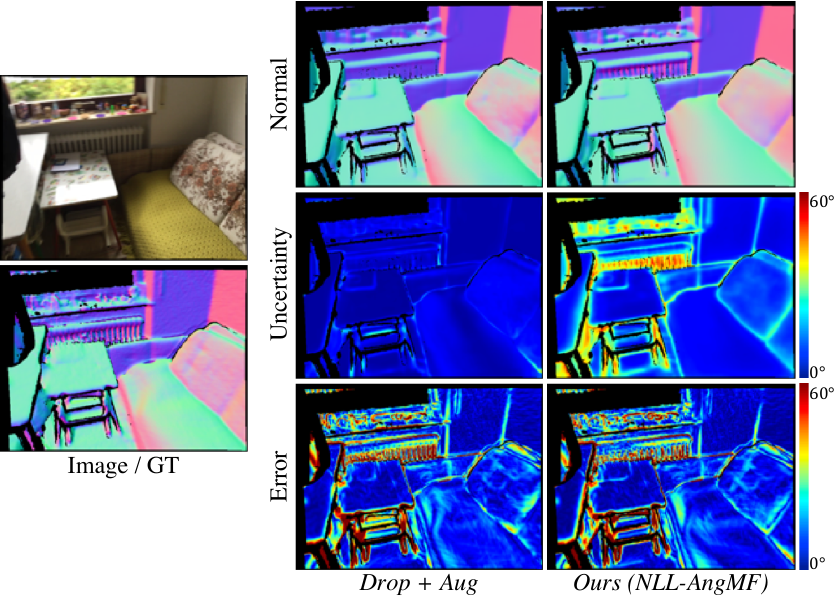}
\end{center}
\caption{We compare the uncertainty estimated by our method against the uncertainty estimated by applying test-time dropout and augmentation (\textit{Drop+Aug}). The uncertainty estimated by our method shows higher correlation with the prediction error.}
\label{fig:uncertainty_comparison}
\end{figure}

Lastly, we evaluate the quality of the estimated uncertainty by plotting the sparsification curves. As no previous work has estimated the surface normal uncertainty, we compare our method against task-independent approaches. \textit{(1) Test-time dropout (Drop):} 2D dropout ($p=0.2$) is added after each 2D convolutional block in decoder. After training, 8 forward passes are performed, with dropout enabled. \textit{(2) Test-time augmentation (Aug):} Following \cite{on_the_uncertainty}, we perform 2 forward passes by flipping the input image. \textit{(3) Combined approach (Drop + Aug):} We apply the image flipping to the network with dropout to make $2 \! \times \! 8 \! = \! 16$ forward passes. For all three methods, the uncertainty is measured as the average angular error with respect to the mean direction. As the uncertainty cannot be estimated in a single forward pass, the uncertainty-guided sampling is disabled, and the networks are trained with the angular loss. Quantitative results in Tab. \ref{table:uncertainty-nyu} and Tab. \ref{table:uncertainty-scannet} show that the proposed method outperforms other methods across all metrics. Fig. \ref{fig:roc} compares the sparsification curves. When evaluated on all pixels, all methods perform similarly. However, as the pixels with high uncertainty are removed, our method gets significantly more accurate than the others. This suggests that our uncertainty correlates better with the prediction error (see Fig. \ref{fig:uncertainty_comparison} for qualitative comparison).  

\subsection{Supplementary Material}

In the supplementary material, we provide the derivations for the AngMF distribution, quantitative evaluation with additional metrics, cross-dataset evaluation on KITTI \cite{KITTI} and DAVIS \cite{DAVIS} and discussion on failure modes.

\section{Conclusion}

In this work, we estimated and evaluated the aleatoric uncertainty in CNN-based surface normal estimation, for the first time in literature. The proposed method estimates the per-pixel surface normal probability distribution, from which the expected angular error can be inferred to quantify the aleatoric uncertainty. We introduced a new parameterization for the surface normal probability distribution, such that its negative log-likelihood is the angular loss with learned attenuation. We also proposed a novel decoder framework where pixel-wise MLPs are trained on a subset of pixels selected based on the uncertainty. Such uncertainty-guided sampling prevents the bias in training towards large planar surfaces, thereby improving the level of detail in the prediction. Experimental results show that the proposed method achieves state-of-the-art performance on ScanNet \cite{ScanNet} and NYUv2 \cite{NYUv2}, and that the estimated uncertainty correlates well with the prediction error.

{\small
\bibliographystyle{ieee_fullname}
\bibliography{egbib}
}

\appendix
\onecolumn

\noindent
\textit{\textbf{\large Supplementary material}}

\noindent
In this supplementary material, we provide (\ref{sec:derivation}) the derivations for the proposed AngMF distribution, (\ref{sec:quantitative}) quantitative evaluation with additional metrics, (\ref{sec:generalization}) cross-dataset evaluation on KITTI \cite{KITTI} and DAVIS \cite{DAVIS}, (\ref{sec:failure}) discussion on failure modes and (\ref{sec:qualitative}) additional qualitative comparison against the state-of-the-art.

\section{Derivations for the proposed AngMF distribution}
\label{sec:derivation}

In the paper, we introduced a variant of the von Mises-Fisher distribution \cite{fisher1993statistical}, such that its negative log-likelihood (NLL) is the angular loss with learned attenuation. We call this the Angular vonMF (AngMF) distribution. In this section, we provide the derivations for Eq. \ref{eqn:ang-PDF}, Eq. \ref{eqn:ang-NLL} and Eq. \ref{eqn:ang-uncertainty}.

\subsection{Probability density function (Eq. \ref{eqn:ang-PDF})}

\begin{figure*}[t]
\begin{center}
\includegraphics[width=0.9\linewidth]{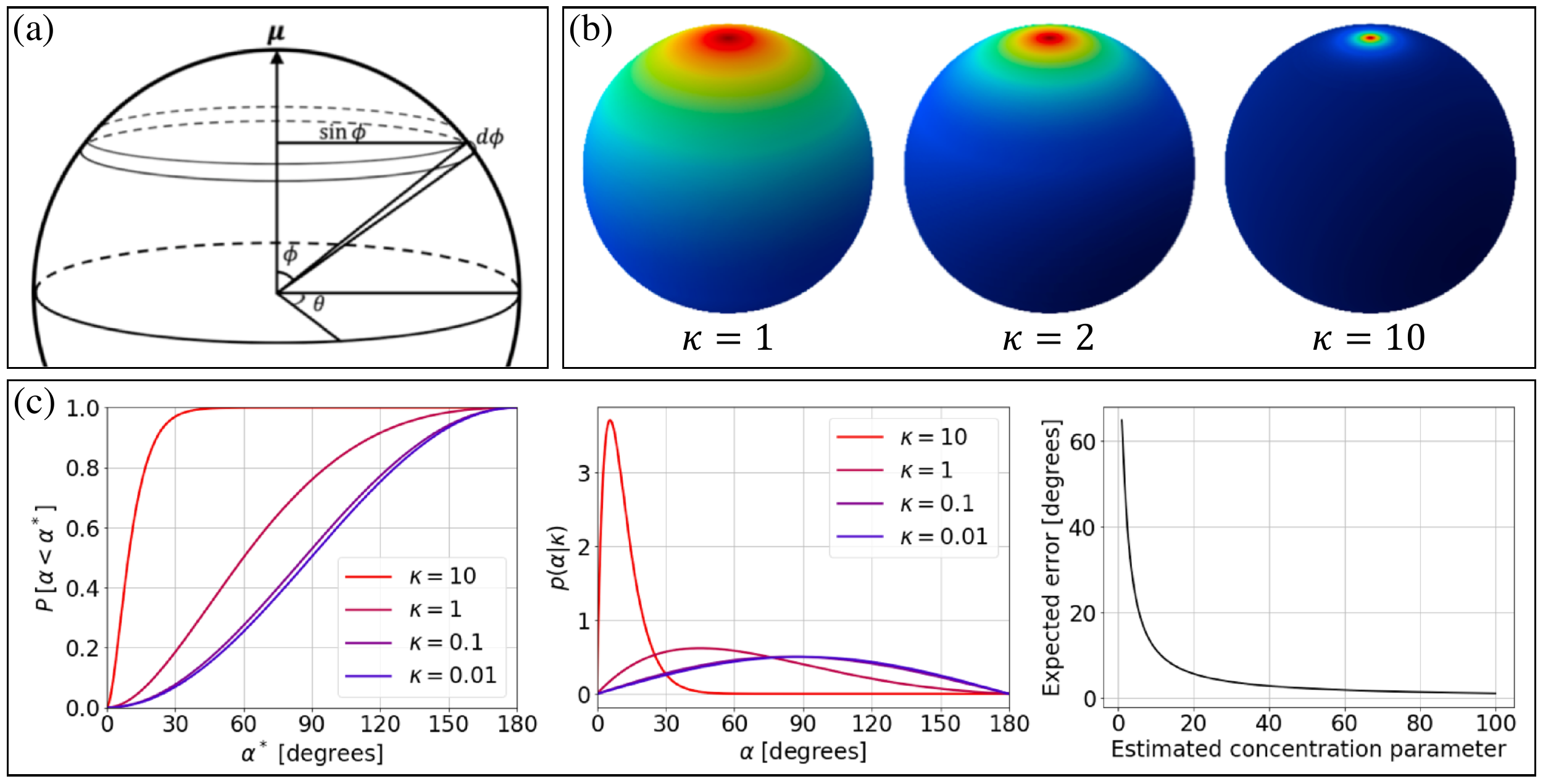}
\end{center}
\caption{(a) The axes orientation used for the integrations in Eq. \ref{eqn:supp_new_cdf} and Eq. \ref{eqn:supp_exp_error}. The mean direction $\boldsymbol{\mu}$ is aligned with the $z$-axis, and is thus excluded in the integration. (b) Visualization of Eq. \ref{eqn:supp_new_pdf2} for different values of $\kappa$. $\kappa$ determines how concentrated the distribution is towards the mean direction. (c) Eq. \ref{eqn:supp_ours_cdf2}, Eq. \ref{eqn:supp_alpha_pdf} and Eq. \ref{eqn:supp_exp_error} plotted for different values of $\kappa$. The expected error decreases as the confidence $\kappa$ increases.}
\label{fig:supp_dist_vis}
\end{figure*}

The NLL of the distribution should have the form of

\begin{equation}
\begin{aligned}
\label{eqn:supp_new_nll}
\mathcal{L}_i = C(\kappa_i) + \kappa_i \cos^{-1} \boldsymbol{\mu}_i^T \mathbf{n}_i^{gt},
\end{aligned}
\end{equation}

\noindent
where $i$ is the pixel index and $\cos^{-1} \boldsymbol{\mu}_i^T \mathbf{n}_i^{gt}$ is the angle between the predicted mean direction $\boldsymbol{\mu}_i$ and the ground truth surface normal $\mathbf{n}_i^{gt}$. The angular error is weighted by the concentration parameter $\kappa_i$, which encodes the network's confidence in the predicted mean direction. The first term $C(\kappa_i)$ should be a monotonically decreasing function of $\kappa_i$ in order to prevent the trivial solution where $\kappa_i = 0 \; \forall \; i$. Then, the probability density function (PDF) should look like

\begin{equation}
\begin{aligned}
\label{eqn:supp_new_pdf}
p(\mathbf{n}_i|\boldsymbol{\mu}_i,\kappa_i)
= D(\kappa_i) \exp (-\kappa_i \cos^{-1} 
\boldsymbol{\mu}_i^T \mathbf{n}_i),
\end{aligned}
\end{equation}

\noindent
where $C(\kappa_i) = -\log(D(\kappa_i))$. We can then compute the cumulative probability of the angular error $\cos^{-1} 
\boldsymbol{\mu}_i^T \mathbf{n}_i$ being less than some threshold $\alpha^*$. See Fig. \ref{fig:supp_dist_vis}-(a) for the axes orientation used for the integration.
 
\begin{equation}
\begin{aligned}
\label{eqn:supp_new_cdf}
P[\cos^{-1} (\boldsymbol{\mu}_i^T \mathbf{n}_i) \leq \alpha^*]
&= 
\int^{2\pi}_{0} 
\int^{\alpha^*}_0 
D(\kappa_i)
\exp(-\kappa_i \phi) \sin \phi d\phi d\theta
\\
&= 2\pi D(\kappa_i)
\int^{\alpha^*}_0
\exp(-\kappa_i \phi) \sin \phi d\phi 
\\
&= 2\pi D(\kappa_i) 
\bigg[ \frac{-\exp(-\kappa_i \phi) (\cos \phi + \kappa_i \sin \phi)}{\kappa_i^2 + 1} \bigg]^{\alpha^*}_0 
\\
&= 2\pi D(\kappa_i)
\frac{1 - \exp(-\kappa_i \alpha^*) (\cos \alpha^* + \kappa_i \sin \alpha^*)}{\kappa_i^2 + 1}
\end{aligned}
\end{equation}

\noindent
Solving $P[\cos^{-1} (\boldsymbol{\mu}_i^T \mathbf{n}_i) \leq \pi] = 1$ gives

\begin{equation}
\begin{aligned}
\label{eqn:supp_ours_D}
D(\kappa_i) = \frac{1}{2\pi} \frac{\kappa_i^2 + 1}{1 + \exp(-\kappa_i \pi)}.
\end{aligned}
\end{equation}

\noindent
Inserting Eq. \ref{eqn:supp_ours_D} into Eq. \ref{eqn:supp_new_pdf} gives 

\begin{equation}
\begin{aligned}
\label{eqn:supp_new_pdf2}
p_{i}(\mathbf{n}_i|\boldsymbol{\mu}_i,\kappa_i) = 
\frac{(\kappa_i^2+1)\exp(-\kappa_i \cos^{-1} \boldsymbol{\mu}_i^T \mathbf{n}_i)}{2\pi (1 + \exp(-\kappa_i \pi))},
\end{aligned}
\end{equation}

\noindent
which is Eq. \ref{eqn:ang-PDF}. Fig. \ref{fig:supp_dist_vis}-(b) visualizes the distribution for different values of $\kappa$. As $\kappa$ increases, the distribution becomes more concentrated around the mean direction. 

\subsection{Negative log-likelihood (Eq. \ref{eqn:ang-NLL})}

The network is trained by minimizing the NLL of the ground truth normal. The training loss can thus be written as

\begin{equation}
\begin{aligned}
\label{eqn:supp_new_nll2}
\mathcal{L}_i = - \log (\kappa_i^2 + 1)
+ \log (1 + \exp(-\kappa_i \pi))
+ \kappa_i \cos^{-1} \boldsymbol{\mu}_i^T \mathbf{n}^\text{gt}_i,
\end{aligned}
\end{equation}

\noindent
where we drop the constant term, $\log 2\pi$. This is Eq. \ref{eqn:ang-NLL} in the paper.

\subsection{Measure of uncertainty (Eq. \ref{eqn:ang-uncertainty})}

Inserting Eq. \ref{eqn:supp_ours_D} to Eq. \ref{eqn:supp_new_cdf} gives

\begin{equation}
\label{eqn:supp_ours_cdf2}
P[\cos^{-1} (\boldsymbol{\mu}_i^T \mathbf{n}_i) \leq \alpha^*]
= 
\frac{1 - \exp(-\kappa_i \alpha^*) (\cos \alpha^* + \kappa_i \sin \alpha^*)}{1 + \exp(-\kappa_i \pi)}.
\end{equation}

\noindent
From this, we can calculate the probability density function for the angular error $\alpha$ via differentiation.

\begin{equation}
\begin{aligned}
\label{eqn:supp_alpha_pdf}
p(\alpha|\boldsymbol{\mu}_i,\kappa_i)
&= \frac{d}{d\alpha}
\left(
\frac{1 - \exp(-\kappa_i \alpha) (\cos \alpha + \kappa_i \sin \alpha)}{1 + \exp(-\kappa_i \pi)}
\right) \\
&= 
\frac{- \exp(-\kappa_i \alpha) (-\sin \alpha + \kappa_i \cos \alpha)
+ \kappa_i \exp(-\kappa_i \alpha) (\cos \alpha + \kappa_i \sin \alpha)}
{1 + \exp(-\kappa_i \pi)} \\
&=
\frac{\exp(-\kappa_i \alpha) 
\sin (\alpha)
(\kappa_i^2 + 1)}
{1 + \exp(-\kappa_i \pi)}
\end{aligned}
\end{equation}

\noindent
Then, the expected value of $\alpha$ can be obtained as

\begin{equation}
\begin{aligned}
\label{eqn:supp_exp_error}
E[\alpha]
&=
\int^\pi_0
\alpha
\frac{\exp(-\kappa_i \alpha) 
\sin (\alpha)
(\kappa_i^2 + 1)}
{1 + \exp(-\kappa_i \pi)}
d\alpha \\
&= 
\frac{\kappa_i^2 + 1}{1 + \exp(-\kappa_i \pi)}
\int^\pi_0
\alpha
\exp(-\kappa_i \alpha)
\sin \alpha d\alpha \\
&= \frac{\kappa_i^2 + 1}{1 + \exp(-\kappa_i \pi)}
\Bigg[ -
\frac{
\exp(-\kappa_i \alpha)
((\kappa_i ((\kappa_i^2 + 1) \alpha + \kappa_i) - 1)
\sin \alpha
+ ((\kappa_i^2 + 1) \alpha + 2 \kappa_i) \cos \alpha)
}{
(\kappa_i^2 + 1)^2
}
\Bigg]^\pi_0 \\
&= \frac{\kappa_i^2 + 1}{1 + \exp(-\kappa_i \pi)}
\Bigg[
\frac{
2\kappa_i (1 + \exp(-\kappa_i \pi)) + \exp(-\kappa_i \pi)(\kappa_i^2+1)\pi
}
{(\kappa_i^2 + 1)^2}
\Bigg] \\
&=
\frac{2\kappa_i}{\kappa_i^2+1}
+ \frac{\exp(-\kappa_i \pi)\pi}{1 + \exp(-\kappa_i \pi)},
\end{aligned}
\end{equation}

\noindent
which is Eq. \ref{eqn:ang-uncertainty}. This quantity is used as a measure of the aleatoric uncertainty. Fig. \ref{fig:supp_dist_vis}-(c) visualizes Eq. \ref{eqn:supp_ours_cdf2}, Eq. \ref{eqn:supp_alpha_pdf} and Eq. \ref{eqn:supp_exp_error} for different values of $\kappa$. The expected error decreases as $\kappa$ increases. For $\kappa=0$, the distribution is uniform and the expected error is $\pi/2$.

\section{Quantitative evaluation with additional metrics}
\label{sec:quantitative}

In this section, we provide the quantitative evaluation of our method with additional metrics. Tab. \ref{table:supp_tiltedsn_comparison}, Tab. \ref{table:supp_uncertainty-nyu} and Tab. \ref{table:supp_uncertainty-scannet} are extensions of Tab. \ref{table:BM-scannet}, Tab. \ref{table:uncertainty-nyu} and Tab. \ref{table:uncertainty-scannet}, respectively. Fig. \ref{fig:supp_uncertainty_nyu} and Fig. \ref{fig:supp_uncertainty_scannet} are extensions of Fig. \ref{fig:roc}.

\subsection{Comparison against TiltedSN}

Tab. \ref{table:supp_tiltedsn_comparison} provides comparison against TiltedSN \cite{SNfromRGB_20_TiltedSN} on ScanNet \cite{ScanNet} with additional metrics. Note that the difference in the accuracy (\% of pixels with error less than $t^\circ$) increases for lower thresholds. 

\begin{table}[h]
\normalsize
\begin{center}
\begin{tabular}{l|ccc|ccccc}
\toprule
Method & mean & median & rmse & $5.0^{\circ}$ & $7.5^{\circ}$ & $11.25^{\circ}$ & $22.5^{\circ}$ & $30^{\circ}$ \\
\midrule
TiltedSN\cite{SNfromRGB_20_TiltedSN} 
& 12.6 & 6.0 & 21.1 
& 42.8 & 57.5 & 69.3 & 83.9 & 88.6 \\
Ours 
& \textbf{11.8} & \textbf{5.7} & \textbf{20.0} 
& \textbf{45.1} & \textbf{59.6} & \textbf{71.1} & \textbf{85.4} & \textbf{89.8} \\
\hline
Difference
& -0.8 & -0.3 & -1.1
& +2.3 & +2.1 & +1.8 & +1.5 & +1.2 \\
\bottomrule
\end{tabular}
\end{center}
\caption{Quantitative comparison against TiltedSN \cite{SNfromRGB_20_TiltedSN} on ScanNet \cite{ScanNet}.}
\label{table:supp_tiltedsn_comparison}
\end{table}

\subsection{Quality of the estimated uncertainty}

Tab. \ref{table:supp_uncertainty-nyu} and Tab. \ref{table:supp_uncertainty-scannet} compare different methods of estimating the surface normal uncertainty. "\textit{Drop}" (making 8 inferences with dropout enabled), "\textit{Aug}" (making 2 inferences by flipping the image) and "\textit{Drop+Aug}" (making 8$\times$2 inferences by applying both) are task-independent approaches which do not require the output to be distributional. The proposed pipeline, trained with the NLL losses, significantly outperforms other approaches across all metrics, suggesting that the estimated uncertainty better correlates with the prediction error.

\begin{table}[h]
\normalsize
\setlength\tabcolsep{1.5pt}
\begin{center}
\begin{tabular}{l|cccccc|cccccc}
\toprule
\multirow{2}{4em}{Method} & \multicolumn{6}{c|}{AUSC $\downarrow$} &
\multicolumn{6}{c}{AUSE $\downarrow$}\\
\cline{2-13}
& {\small mean} & {\small median} & {\small rmse} & {\footnotesize $ 11.25^{\circ}$} & {\footnotesize $ 22.5^{\circ}$} & {\footnotesize $ 30.0^{\circ}$} & {\small mean} & {\small median} & {\small rmse} & {\footnotesize $11.25^{\circ}$} & {\footnotesize $22.5^{\circ}$} & {\footnotesize $30.0^{\circ}$}\\
\midrule
\textit{Drop}
& 9.01 & 4.91 & 15.84 & 19.32 & 8.66 & 6.07
& 4.02 & 0.91 & 9.61 & 10.23 & 6.10 & 4.76\\
\textit{Aug} 
& 8.64 & 4.68 & 15.08 & 18.75 & 8.26 & 5.64
& 3.93 & 0.97 & 9.14 & 10.25 & 5.84 & 4.42\\
\textit{Drop + Aug} 
& 8.16 & 4.68 & 14.32 & 16.73 & 7.18 & 4.97
& 3.22 & 0.73 & 8.15 & 7.75 & 4.65 & 3.68\\
\hline
Ours \textit{(NLL-vonMF)} 
& 7.03 & 4.47 & 10.96 & 14.24 & 5.51 & 3.53
& \textbf{2.11} & \textbf{0.56} & \textbf{4.80} & 5.10 & 2.92 & 2.24\\
Ours \textit{(NLL-AngMF)} 
& \textbf{6.83} & \textbf{4.25} & \textbf{10.92} & \textbf{13.47} & \textbf{5.27} & \textbf{3.45}
& 2.13 & \textbf{0.56} & 4.98 & \textbf{5.01} & \textbf{2.86} & \textbf{2.22}\\
\bottomrule
\end{tabular}
\end{center}
\caption{Quantitative evaluation of uncertainty on NYUv2 \cite{NYUv2}.}
\label{table:supp_uncertainty-nyu}
\end{table}

\begin{table}[h]
\normalsize
\setlength\tabcolsep{1.5pt}
\begin{center}
\begin{tabular}{l|cccccc|cccccc}
\toprule
\multirow{2}{4em}{Method} & \multicolumn{6}{c|}{\small AUSC $\downarrow$} &
\multicolumn{6}{c}{\small AUSE $\downarrow$}\\
\cline{2-13}
& {\small mean} & {\small median} & {\small rmse} & {\footnotesize $11.25^{\circ}$} & {\footnotesize $22.5^{\circ}$} & {\footnotesize $30.0^{\circ}$} & {\small mean} & {\small median} & {\small rmse} & {\footnotesize $11.25^{\circ}$} & {\footnotesize $22.5^{\circ}$} & {\footnotesize $30.0^{\circ}$}\\
\midrule
\textit{Drop} 
& 7.25 & 4.35 & 12.51 & 13.95 & 5.49 & 3.60
& 3.24 & 1.02 & 7.55 & 8.58 & 4.14 & 2.94\\
\textit{Aug} 
& 7.06 & 4.08 & 12.58 & 13.72 & 5.36 & 3.48
& 3.32 & 1.03 & 7.92 & 8.81 & 4.13 & 2.87\\
\textit{Drop + Aug} 
& 6.87 & 4.17 & 12.07 & 12.73 & 4.82 & 3.13
& 2.93 & 0.92 & 7.20 & 7.49 & 3.51 & 2.49\\
\hline
Ours \textit{(NLL-vonMF)} 
& 5.84 & 3.92 & 9.30 & 10.31 & 3.21 & 1.94
& \textbf{1.85} & \textbf{0.64} & \textbf{4.38} & 4.69 & \textbf{1.86} & 1.30\\
Ours \textit{(NLL-AngMF)} 
& \textbf{5.64} & \textbf{3.73} & \textbf{9.07} & \textbf{9.48} & \textbf{3.11} & \textbf{1.90}
& 1.88 & 0.66 & \textbf{4.38} & \textbf{4.47} & 1.88 & \textbf{1.29}\\
\bottomrule
\end{tabular}
\end{center}
\caption{Quantitative evaluation of uncertainty on ScanNet \cite{ScanNet}.}
\label{table:supp_uncertainty-scannet}
\end{table}

\newpage
\subsection{Sparsification curves}

\begin{figure}[h]
\begin{center}
\includegraphics[width=1.0\linewidth]{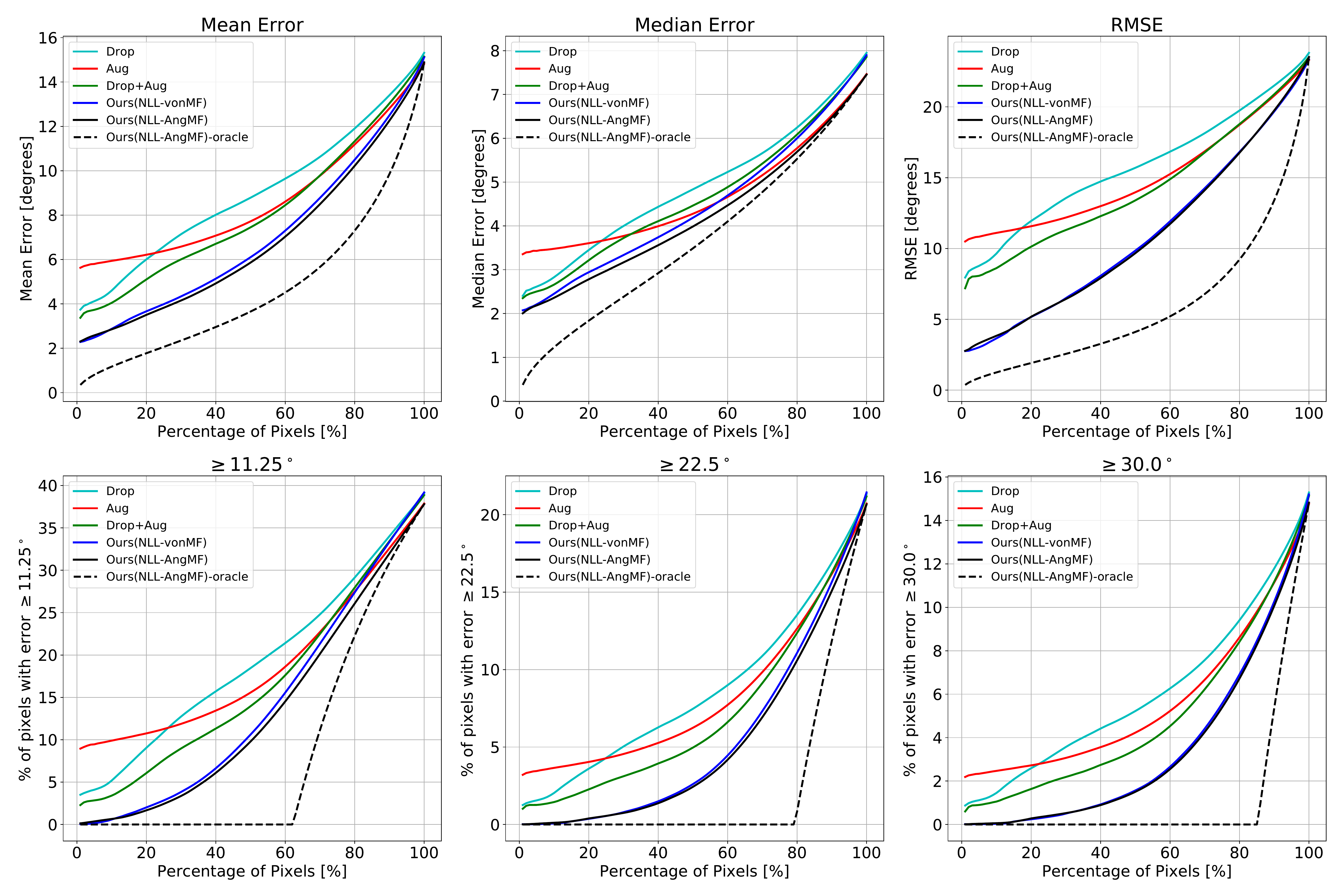}
\end{center}
\caption{Sparsification curves for NYUv2 \cite{NYUv2}.}
\label{fig:supp_uncertainty_nyu}
\end{figure}

\begin{figure}[t]
\begin{center}
\includegraphics[width=1.0\linewidth]{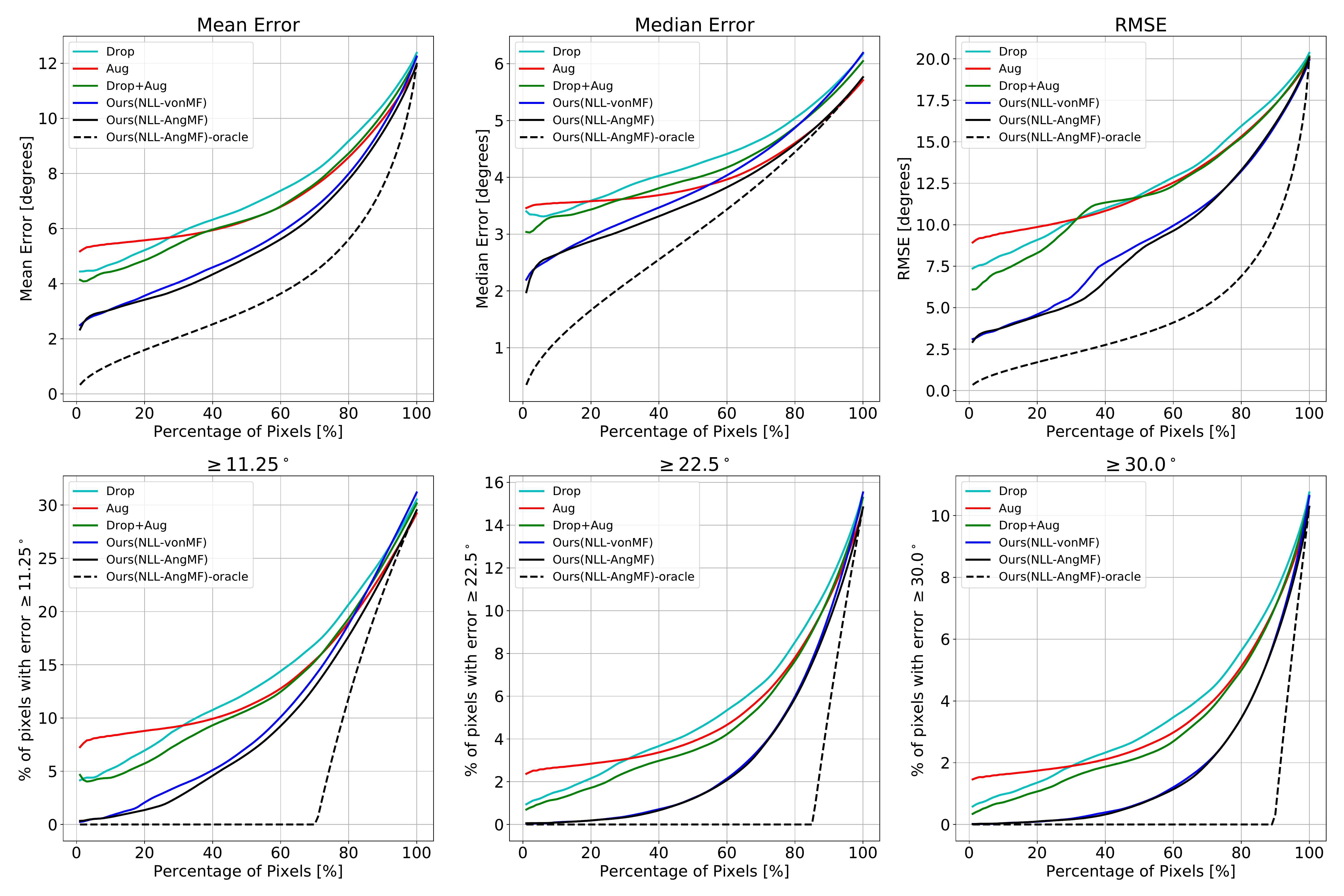}
\end{center}
\caption{Sparsification curves for ScanNet \cite{ScanNet}.}
\label{fig:supp_uncertainty_scannet}
\end{figure}

Fig. \ref{fig:supp_uncertainty_nyu} and Fig. \ref{fig:supp_uncertainty_scannet} provide the sparsification curves for NYUv2 \cite{NYUv2} and ScanNet \cite{ScanNet}, respectively. When evaluated on all pixels, all methods perform similarly. However, as the pixels with high uncertainty are removed, our method gets significantly more accurate than the others, suggesting that our uncertainty correlates better with the prediction error. For "Ours (\textit{NLL-AngMF})", we also show the ideal sparsification (oracle) by sorting the pixels by the error.

\section{Cross-dataset evaluation on KITTI and DAVIS}
\label{sec:generalization}

In the paper, we performed a cross-dataset evaluation by training the network on ScanNet \cite{ScanNet} and testing it on NYUv2 \cite{NYUv2} without fine-tuning. However, this is not a challenging task as both datasets contain images of indoor scenes with similar visual features. In this section, we further demonstrate the generalization ability of our method by testing the network (trained only on ScanNet) on two challenging datasets - KITTI \cite{KITTI} and DAVIS \cite{DAVIS}. The results are provided in Fig. \ref{fig:supp_generalize_kitti} and Fig. \ref{fig:supp_generalize_davis}. For comparison, we also provide the predictions made by TiltedSN \cite{SNfromRGB_20_TiltedSN}.

The ground truth surface normal for ScanNet is calculated from a 3D mesh that is obtained by fusing thousands of depth-maps. For this reason, the ground truth generally does not exist for dynamic objects such as humans. As the dataset is collected in indoor scenes, it also does not contain instances of cars and buildings. Nonetheless, the network can generalize well for such unseen objects. We believe that this is because the network utilizes low-level features, such as edges and shades, which are universal in most datasets. Fig. \ref{fig:supp_fail_layout}, which will be discussed in Sec. \ref{sec:failure}, supports such argument. Even when the input image only contains edges or shades, the network can infer the 3D structure.

\afterpage{
\begin{figure}[p]
\begin{center}
\includegraphics[width=0.8\linewidth]{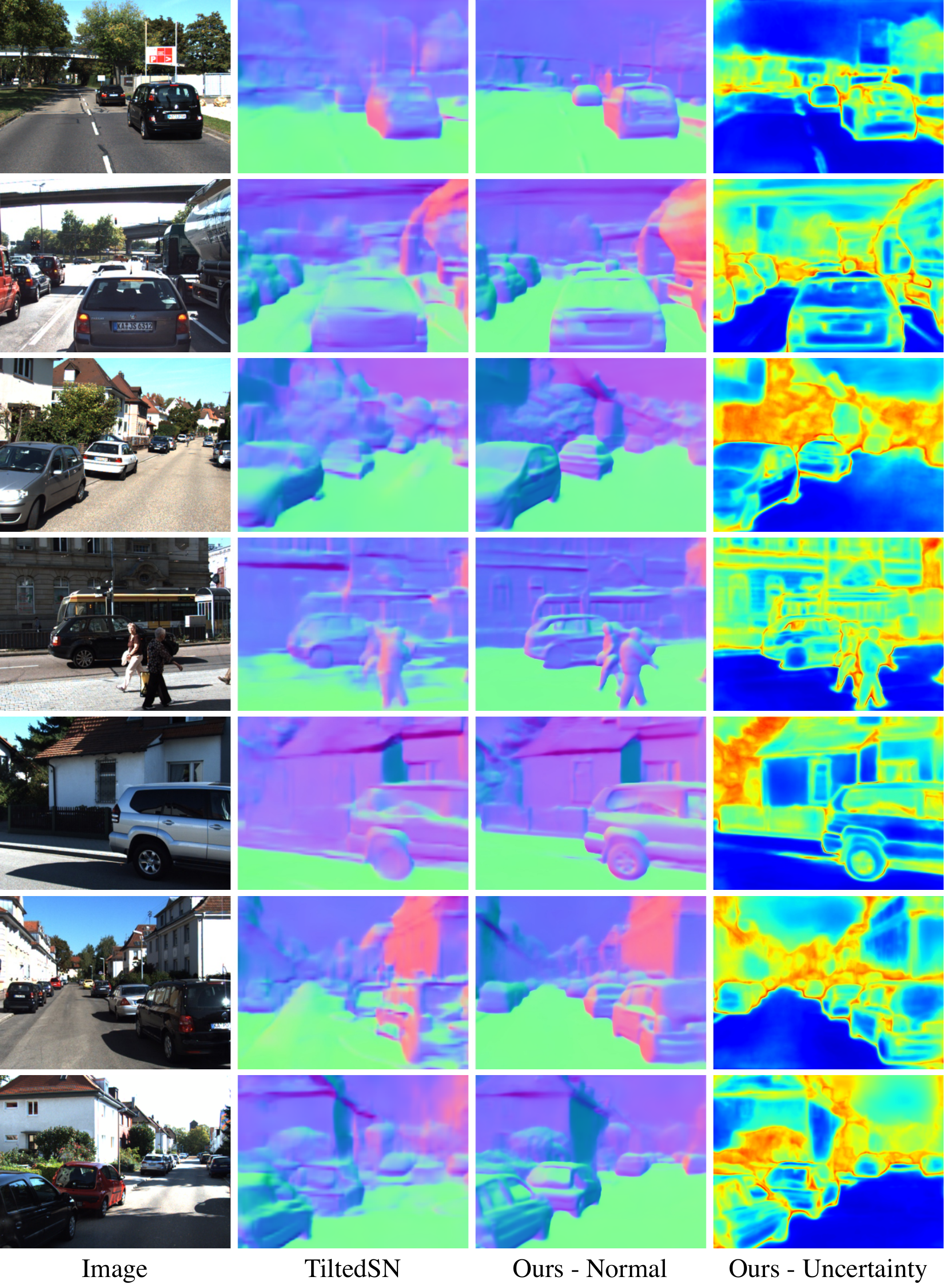}
\end{center}
\caption{Cross-dataset evaluation on KITTI \cite{KITTI}.}
\label{fig:supp_generalize_kitti}
\end{figure}
\clearpage}

\afterpage{
\begin{figure}[p]
\begin{center}
\includegraphics[width=0.8\linewidth]{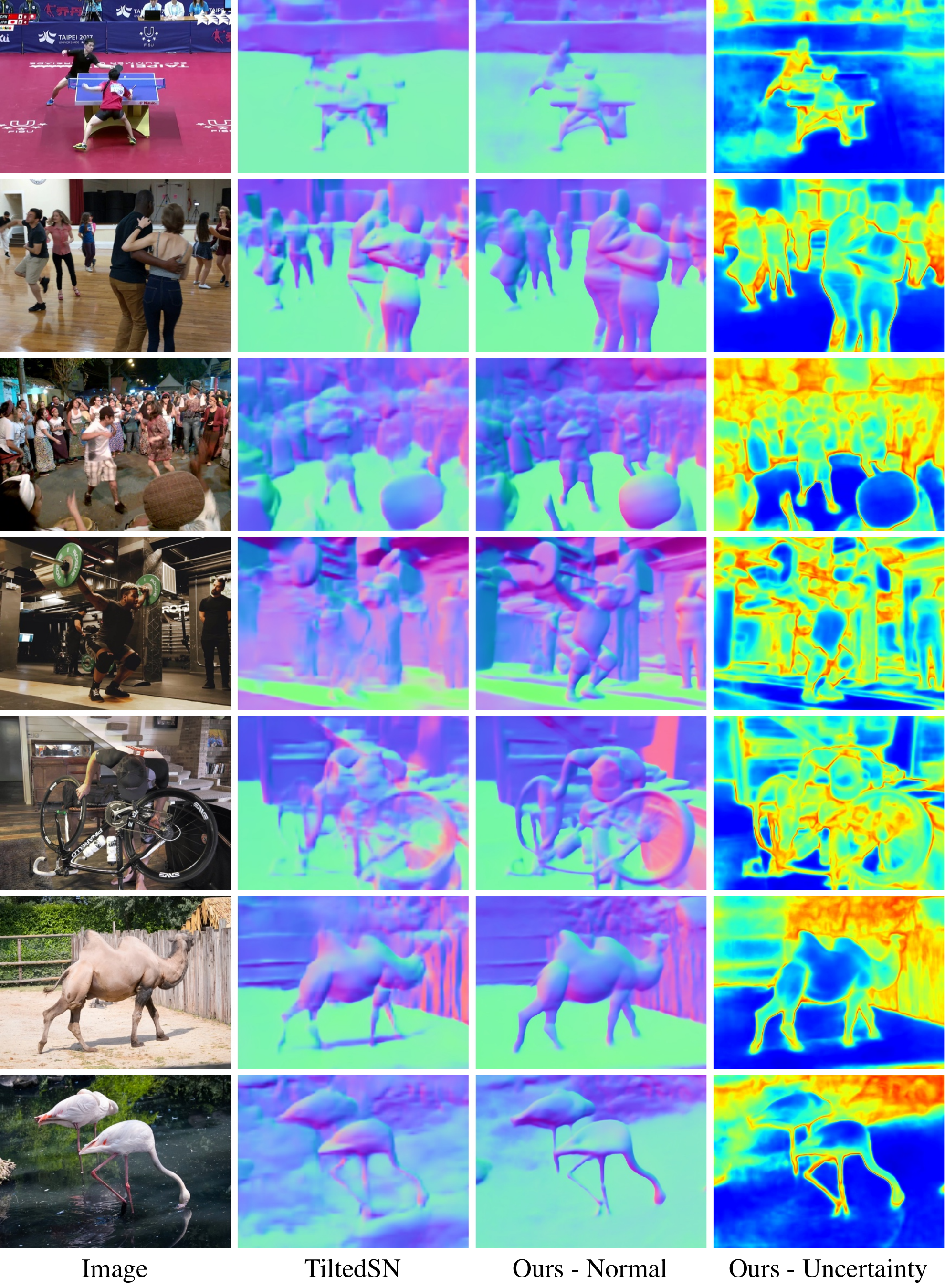}
\end{center}
\caption{Cross-dataset evaluation on DAVIS \cite{DAVIS}.}
\label{fig:supp_generalize_davis}
\end{figure}
\clearpage}

\newpage
\section{Failure modes}
\label{sec:failure}

In this section, we discuss the failure modes of the proposed method. 

\subsection{Tilted images}

\begin{figure}[h]
\begin{center}
\includegraphics[width=0.95\linewidth]{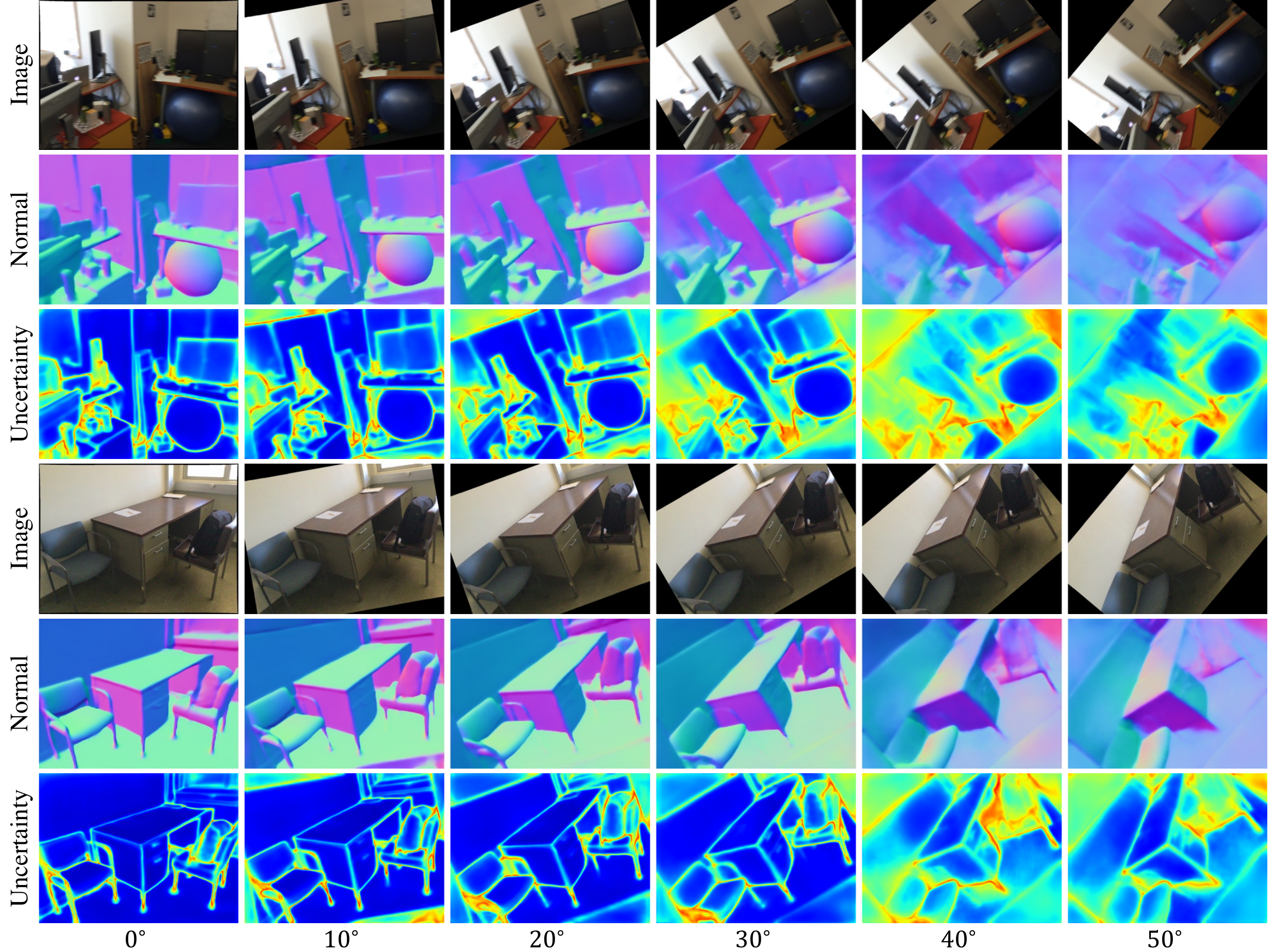}
\end{center}
\caption{Predictions made for tilted images.}
\label{fig:supp_fail_rotation}
\end{figure}

Fig. \ref{fig:supp_fail_rotation} shows the predictions made for tilted images. The network is robust against mild rotations ($\sim \!\! 20^\circ$), but suffers when the images are tilted severely ($30^\circ \!\! \sim$). Nevertheless, the expected error (clamped between $0^\circ$ and $60^\circ$ in all images) also increases for such images, demonstrating the usefulness of the estimated uncertainty. Tilted images can be handled by using a spatial rectifier to warp the images such that its surface normal distribution matches to that of the training images, as done in \cite{SNfromRGB_20_TiltedSN}. This will be investigated in our future work.

\subsection{Inherent ambiguity of the problem}

\begin{figure}[h]
\begin{center}
\includegraphics[width=0.95\linewidth]{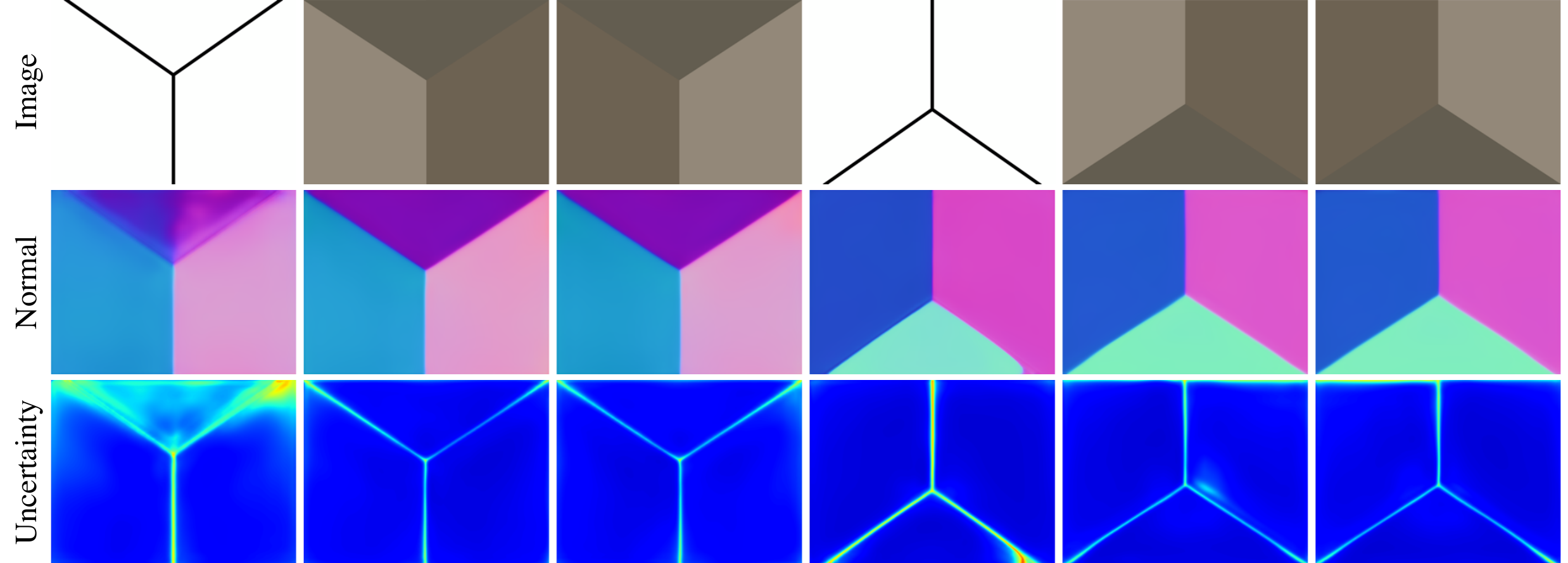}
\end{center}
\caption{Predictions made for artificial images consisting only of edges and shades.}
\label{fig:supp_fail_layout}
\end{figure}

To understand the visual cues used by the network, we created artificial images consisting only of edges and shades. Fig. \ref{fig:supp_fail_layout} shows the predictions made by the network. The first three images show "Y"-shaped structures and the other three are their vertically flipped versions. Note that the depth of each pixel can have any arbitrary value, meaning that the surface can have any form. It can be a concave (or convex) corner or even a drawing on a flat wall.

For the last three images, the network thinks that it is a concave corner. This is because such structure was mostly seen in the lower corners of cuboid-shaped rooms. However, the prediction is not clear for the "Y"-shaped structures. We believe that this is because such structure was seen in both concave corners (upper corners of rooms) and convex corners (external corners of furnitures). To handle such ambiguity, the network should estimate a \textit{multi-modal} surface normal distribution, which consists of multiple uni-modal distributions with mixing coefficients. This will be investigated in our future work.

\section{Additional comparison against the state-of-the-art}
\label{sec:qualitative}

Lastly, we provide additional qualitative comparison against GeoNet++ \cite{SNfromRGB_20_GeoNet++} (in Fig. \ref{fig:supp_bm_nyu}) and TiltedSN \cite{SNfromRGB_20_TiltedSN} (in Fig. \ref{fig:supp_bm_scannet}).

\afterpage{
\begin{figure}[p]
\begin{center}
\includegraphics[width=1.0\linewidth]{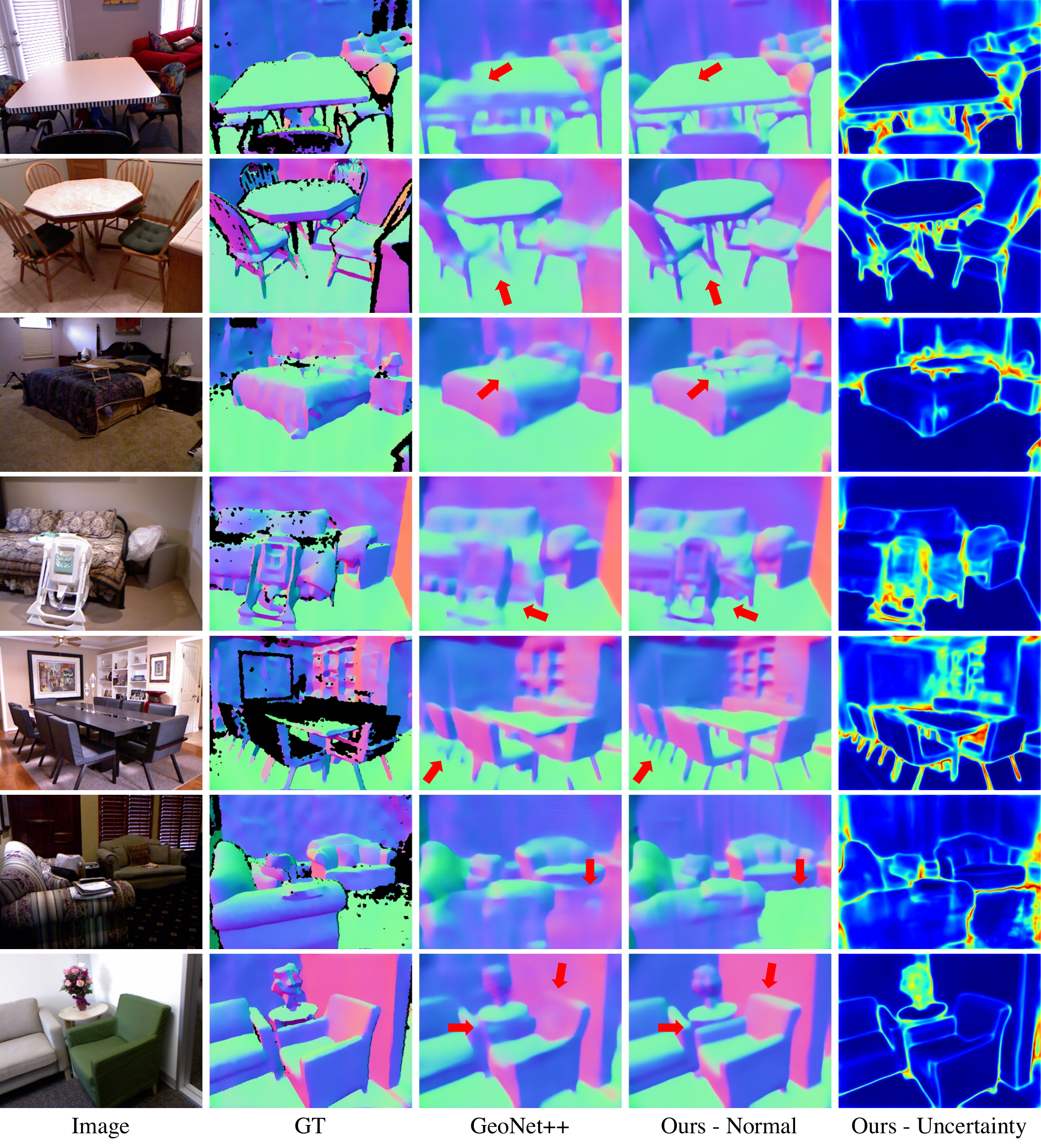}
\end{center}
\caption{Additional qualitative comparison against GeoNet++ \cite{SNfromRGB_20_GeoNet++} on NYUv2 \cite{NYUv2}. Despite the poor quality of the ground truth, our method can recover the fine details of the scene geometry (see the areas pointed by the red arrows).}
\label{fig:supp_bm_nyu}
\end{figure}
}

\afterpage{
\begin{figure}[p]
\begin{center}
\includegraphics[width=1.0\linewidth]{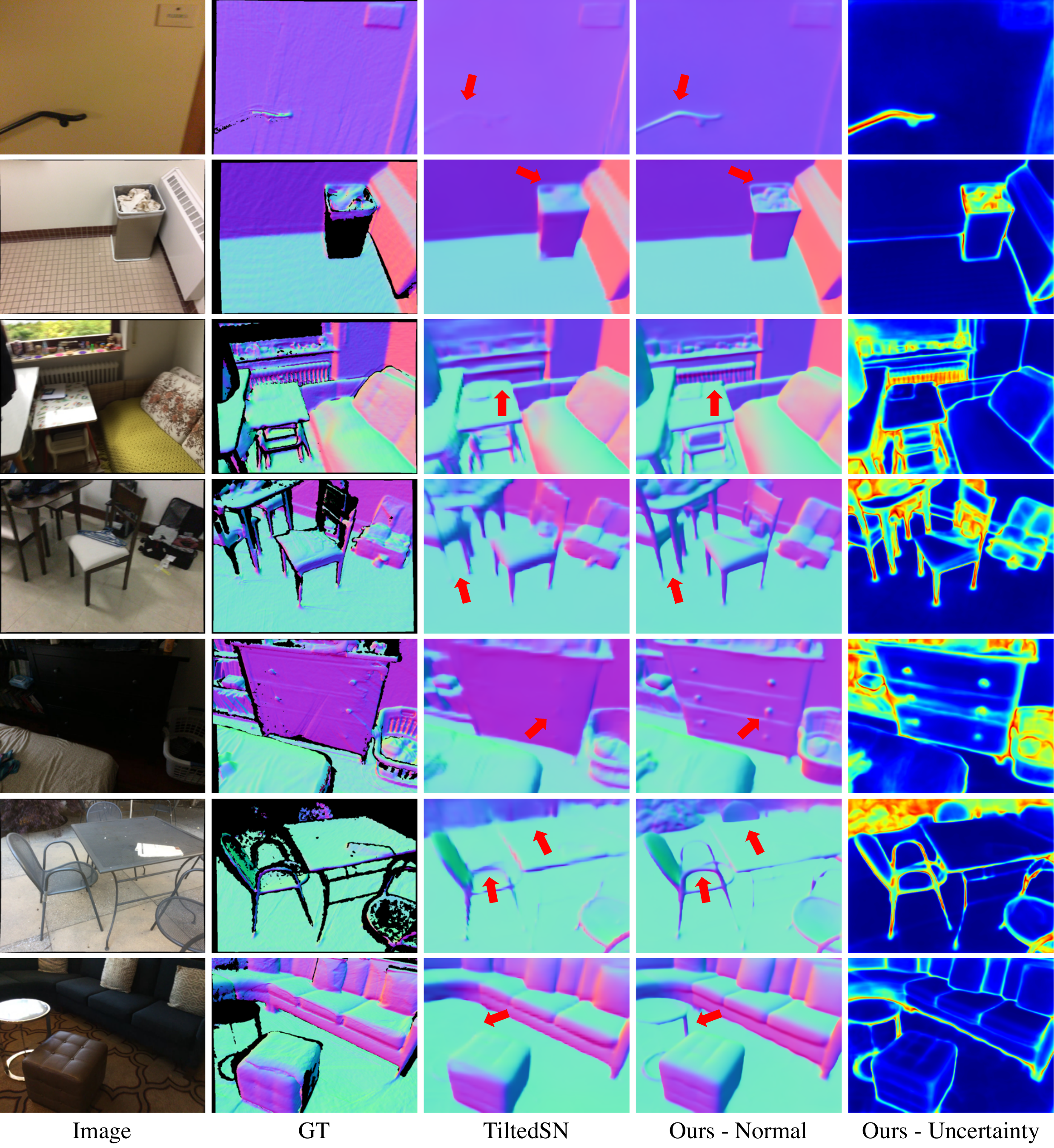}
\end{center}
\caption{Additional qualitative comparison against TiltedSN \cite{SNfromRGB_20_TiltedSN} on ScanNet \cite{ScanNet}. 
The predictions made by our method contain higher level of detail (see the areas pointed by the red arrows).}
\label{fig:supp_bm_scannet}
\end{figure}
}


\end{document}